\documentclass[3p]{elsarticle}

\usepackage{amsmath,amsfonts}
\usepackage{algorithmic}
\usepackage{algorithm}
\usepackage{array}
\usepackage[caption=false,font=scriptsize,labelfont=sf,textfont=sf]{subfig}
\usepackage{textcomp}
\usepackage{stfloats}
\usepackage{url}
\usepackage{verbatim}
\usepackage{graphicx}

\usepackage{color}

\usepackage{algorithmic}
\usepackage{algorithm}

\usepackage[mathscr]{eucal}
\usepackage{amssymb}
\usepackage{subfig} 
\usepackage{subfloat}
\usepackage{multirow}
\usepackage{hyperref}
\usepackage{bm}

\makeatletter
\renewcommand{\fnum@figure}{Fig. \thefigure.\@gobble}
\makeatother

\biboptions{sort&compress}
\journal{Journal of Applied Soft Computing}

\begin{document}

\begin{frontmatter}

\title{Multitasking Evolutionary Algorithm Based on Adaptive Seed Transfer for Combinatorial Problem}
%\tnotetext[mytitlenote]{Fully documented templates are available in the elsarticle package on \href{http://www.ctan.org/tex-archive/macros/latex/contrib/elsarticle}{CTAN}.}

%% Group authors per affiliation:

\author[mymainaddress]{Haoyuan Lv}

\author[mymainaddress]{Ruochen Liu\corref{mycorrespondingauthor}}
\cortext[mycorrespondingauthor]{Corresponding author}
\ead{ruochenliu@xidian.edu.cn}

\address[mymainaddress]{Key Lab of Intelligent Perception and Image Understanding of Ministry of Education, International Center of Intelligent Perception and Computation, Xidian University, Xi'an, 710071, China}

\begin{abstract}
Evolutionary computing (EC) is widely used in dealing with combinatorial optimization problems (COP). Traditional EC methods can only solve a single task in a single run, while real-life scenarios often need to solve multiple COPs simultaneously. In recent years, evolutionary multitasking optimization (EMTO) has become an emerging topic in the EC community. And many methods have been designed to deal with multiple COPs concurrently through exchanging knowledge. However, many-task optimization, cross-domain knowledge transfer, and negative transfer are still significant challenges in this field. A new evolutionary multitasking algorithm based on adaptive seed transfer (MTEA-AST) is developed for multitasking COPs in this work. First, a dimension unification strategy is proposed to unify the dimensions of different tasks. And then, an adaptive task selection strategy is designed to capture the similarity between the target task and other online optimization tasks. The calculated similarity is exploited to select suitable source tasks for the target one and determine the transfer strength. Next, a task transfer strategy is established to select seeds from source tasks and correct unsuitable knowledge in seeds to suppress negative transfer. Finally, the experimental results indicate that MTEA-AST can adaptively transfer knowledge in both same-domain and cross-domain many-task environments. And the proposed method shows competitive performance compared to other state-of-the-art EMTOs in experiments consisting of four COPs.
\end{abstract}


\end{frontmatter}
%-------------------------------------------------------------I.	Introduction------------------------------------------------
\section{Introduction}
Combinatorial optimization problems (COP) widely exist in daily production and life, such as path planning, time allocation, and workshop scheduling. Due to universality and NP-hard characteristic, COP has received extensive attention from many researchers in the field of optimization. Evolutionary Computation (EC) is a class of gradient-free optimization algorithms that are inspired by nature. They are widely used to solve COPs. Traditionally, EC always solves a single problem with a single run. However, there are many real-world scenarios where multiple COPs need to be optimized simultaneously. Some problems represented by multi-scenario optimization focus on solving multiple homogeneous problems \cite{guptaHalfDozenRealWorld2022}. For example, multi-unmanned aerial vehicle (UAV) path planning aims at optimizing the flight trajectories of multiple aircraft at the same time, and vehicle routing problems focus on optimizing the transportation routes of multiple vehicles simultaneously. Another class of problems represented by multi-hard optimization aims to solve multiple heterogeneous problems \cite{guptaBackRootsMultiX2019}. Taking a logistics company as an example, it is necessary to consider the optimization of warehouse location and vehicle routing simultaneously to reduce the company's transportation costs. Furthermore, it is worth noting that these optimization tasks seldom exist in isolation \cite{guptaInsightsTransferOptimization2018, TAN2023110182}. When the EC solver handles multiple COPs, although the individual meanings may vary, some excellent genes still adapt to different tasks. The search efficiency can be accelerated if these genes are shared among different tasks as common knowledge.

In recent years, evolutionary multitasking optimization (EMTO) has become an emerging field in the community of EC. EMTO is widely used to solve various complex optimization problems due to its potential parallelism and excellent search ability. However, most of the knowledge transfer operators in existing EMTO frameworks are designed for continuous optimization with real number coding. They cannot directly extend to COP with discrete encoding.

To explore the feasibility of EMTO to solve COP, many scholars developed a series of studies. Firstly, Gupta \emph{et al.} proposed multifactorial optimization (MFO) \cite{guptaMultifactorialEvolutionEvolutionary2016}. They designed a new multifactorial evolutionary algorithm (MFEA) and used it to deal with the knapsack problem, quadratic assignment problem (QAP), and capacitated vehicle routing problem (CVRP) in multitasking environments. After that, Yuan \emph{et al.} extended MFEA to various permutation-based combinatorial optimization problems, i.e., QAP, traveling salesman problem (TSP), linear ordering problem (LOP), and job-shop scheduling problem (JSP) \cite{yuanEvolutionaryMultitaskingPermutationbased2016}. Zhou \emph{et al.} designed a permutation-based MFEA (P-MFEA) and applied it to solve multitasking CVRP \cite{zhouEvolutionaryMultitaskingCombinatorial2016}. Hao \emph{et al.} proposed an evolutionary multitasking graph-based hyper-heuristic (EMHH) and used it to solve coloring problems and exam timetable problems \cite{haoUnifiedFrameworkGraphBased2021}. Feng \emph{et al.} extended EMTO methods by explicit autoencoding and applied it to deal with multitasking CVRP \cite{fengExplicitEvolutionaryMultitasking2021}. In addition, by combining MFO with a cellular genetic algorithm (CGA), Osaba \emph{et al.} proposed a multifactorial cellular genetic algorithm (MFCGA) for multitasking TSP \cite{osabaMultifactorialCellularGenetic2020}. They also improved MFEA-II as DMFEA-II to solve TSP and CVRP in many-task environments \cite{osabaDMFEAIIAdaptiveMultifactorial2020}. Adaptive transfer-guided multifactorial cellular genetic algorithm (AT-MFCGA) was developed on the basis of MFCGA to explore the performance of MFO in many-task environments and cross-domain transfer \cite{ osabaATMFCGAAdaptiveTransferguided2021}.

Although many excellent EMTO methods have been proposed to address multitasking COPs, some key issues still remain in this field. First, each target task is affected by both positive and negative tasks in a many-task environment, weakening the effect of positive transfer and leading to negative transfer. Therefore, adaptively capturing the similarity between tasks and dynamically adjusting the transfer strength is the key to solving many-task optimization (MaTO). Most current methods adopt feedback-based task selection strategies, which obtain the parameters of the knowledge mixture model by collecting the interaction information between tasks. However, such methods often suffer from a large number of inefficient transfers or even negative transfers in the stage of exploring task relationships.

Furthermore, cross-domain transfer between heterogeneous problems is also difficult when extending EMTO into COP. The dimensions, individual representations, and fitness calculations are all different. Therefore, the knowledge of individuals responsible for different tasks is various. Blindly performing cross-domain knowledge transfer will lead to the failure of knowledge transfer between different tasks and cause severe negative transfer. Most of the existing MFO methods adopt a unified search space strategy to construct a unified problem representation. However, the redundant dimensions introduced by this mechanism are also one of the most important factors for negative transfer \cite{ wuOrthogonalTransferMultitask2022, CHEN2023110070}.

\begin{figure}[ht]
    \begin{center}
        \centerline{\includegraphics[width=0.97\linewidth]{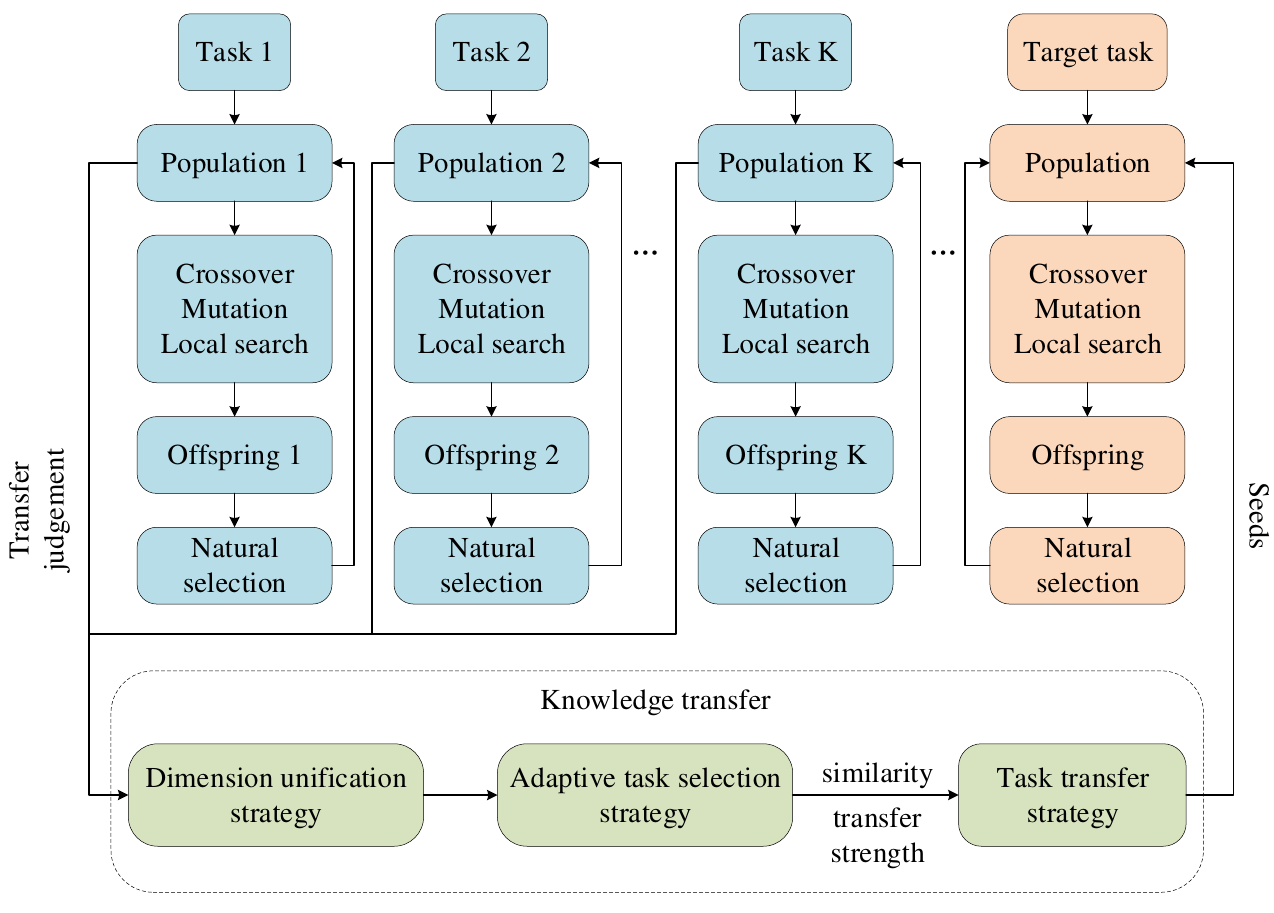}}
        \caption{ The flowchart of MTEA-AST.\label{flowchart}}
    \end{center}
\end{figure}

To address these issues, a new evolutionary multitasking algorithm based on adaptive seed transfer (MTEA-AST) is proposed in this work. As is shown in Fig. \ref{flowchart}, each task in MTEA-AST is assigned a separate population with a specific individual representation and search operator. Dimension unification strategy is used to map individuals to a unified dimension. Then, the adaptive task selection strategy is designed to dynamically calculate the similarity between different tasks and the transfer strength between them. After that, a task transfer strategy is used to select excellent seeds from source populations and transfer them to the population responsible for the target task. Experimental results show that the proposed method achieves competitive performance compared to several EMTO methods and single-task solvers for COP.

The main contributions of MTEA-AST are summarized as follows:

1) To the best known of our knowledge, MTEA-AST is the first explicit transfer method for cross-domain MaTO. In MTEA-AST, a new similarity-based adaptive task selection strategy is designed for COP, which can be applied to a variety of COPs to dynamically capture inter-task similarities and adaptively adjust the transfer intensity. It greatly improves the inefficient transfer and negative transfer caused by the existing feedback-based task selection methods for MaTO in the task relationship exploration stage.

2) In order to overcome the severe negative transfer caused by the inherent differences of heterogeneous problems in the process of cross-domain transfer. Simple but efficient heuristics are used to unify individual representations and suppress negative transfer in the proposed framework. It greatly improves the inefficient transfer caused by redundant encoding in current methods and the knowledge difference of heterogeneous problems.

The rest of this work is organized as follows. Section II introduces the background of EMTO, MaTO, and COP. Some existing EMTO methods and heuristics for solving COP are also presented in this section. Section III describes the details of the proposed algorithm. Experimental results and analysis on same-domain and cross-domain task benchmarks in the many-task environment are presented in Section IV. Section V presents the conclusions and future work.

%-------------------------------------------------------------II.	Background------------------------------------------------
\section{Background}
In this section, the background knowledge of EMTO and existing methods are presented. After that, the definitions of the four COPs involved in this work and some related methods for solving them are discussed.

%---------------------------II.	A.	Related works ---------------------------------
\subsection{Evolutionary multitasking optimization}
Evolutionary multitasking optimization focuses on solving multiple optimization problems concurrently. Generally, an EMTO problem involving $K$ tasks is defined as follows:

\begin{equation}
	\label{eq1}
	x_{i}^{*}=\underset{x\in {{\chi }_{i}}}{\mathop{\arg \min }}\,{{f}_{i}}(x),\text{ }i=1,2,...,K
\end{equation}

Denote ${{T}_{i}}$ as the $i$-th optimization task, where ${{\chi }_{i}}$ and ${{f}_{i}}(.)$ are the search space and objective function of ${{T}_{i}}$, respectively. The goal of EMTO is to find a set ${{X}^{*}}=[x_{1}^{*},x_{2}^{*},...,x_{K}^{*}]$ including the optimal solutions for all tasks by transfer knowledge between them. In the early stage of EMTO, most studies focused on knowledge transfer between a small number of tasks, such as two-task pairs \cite{10040733}. According to \cite{ chenAdaptiveArchiveBasedEvolutionary2020}, many-task optimization is proposed to meet the scenarios where the number of EMTO tasks is more than three. Since then, MaTO has also become an important research field in EMTO.

Due to its potential parallelism and good scalability, the earliest MFO method, i.e., MFEA \cite{guptaMultifactorialEvolutionEvolutionary2016} has been widely studied and applied in the field of EMTO. It adopts a unified search space and implicit transfer strategy to realize knowledge transfer. The former introduces random coding to length low-dimensional task dimensions to the largest dimension of all tasks. And the latter transfer knowledge through probabilistic crossover between individuals responsible for different tasks. Some key definitions are described as follows:

1)	$Factorial\text{ }Cost$: $f_{i}^{j}$ is calculated by the $i$-th individual in the objective function of the $j$-th task.

2)	$Factorial\text{ }Rank$: $r_{i}^{j}$ is the rank position of the $i$-th individual when the population is sorted in ascending order for the $j$-th task.

3)	$Skill\text{ }Factor$: ${{\tau }_{i}}\text{=}\arg {{\min }_{j\in \{1,...,K\}}}r_{i}^{j}$ is the index of the task corresponding to the $i$-th individual.

4)	$Scalar\text{ }Fitness$: ${{\varphi }_{i}}\text{=1/}{{\min }_{j\in \{1,...,K\}}}r_{i}^{j}$ is used for the natural selection operation in the MFEA.

MFO has been applied to solve many traditional optimization problems and real-world problems, such as multi-objective optimization problem \cite{yangMultitaskingMultiobjectiveEvolutionary2019,10032050}, bi-level optimization problem \cite{guptaEvolutionaryMultitaskingBilevel2015}, combinational optimization problem \cite{guptaMultifactorialEvolutionEvolutionary2016, yuanEvolutionaryMultitaskingPermutationbased2016, trungMultifactorialEvolutionaryAlgorithm2019, rauniyarMultifactorialEvolutionaryAlgorithm2019,THANG2021107253}, machine learning \cite{wangEvolutionaryMultitaskingAUC2022, zhongMultifactorialGeneticProgramming2020, nguyenRelatednessMeasuresAid2020, zhangEvolutionaryMachineLearning2022}, unmanned system path planning \cite{ongEvolutionaryMultitaskingComputer2016a, zhouMFEAIGMultiTaskAlgorithm2020, baliCognizantMultitaskingMultiobjective2021}, cloud computing \cite{liuMultiobjectiveMultifactorialEvolutionary2021,10065579} and complex network reconstruction \cite{wuEvolutionaryMultitaskingMultilayer2021,wangLearningLargescaleFuzzy2021b}.

With the development of MFO, some researchers have found negative transfer that hinders the optimization process of the target task during knowledge transfer. Most EMTO frameworks suppress negative transfer from two aspects, task selection, and task transfer \cite{weiReviewEvolutionaryMultiTask2021}. Recently, in \cite{ baliMultifactorialEvolutionaryAlgorithm2020}, Bali \emph{et al.} theoretically proved that fixed random mating probability (RMP) is an important cause of negative transfer and proposed MFEA-II to replace RMP with an online updated RMP matrix. Shakeri \emph{et al.} applied an adversarial multi-arm bandit method to calculate the selection probabilities of each source task \cite{shakeriCopingBigData2019}. Lin \emph{et al.} used an incremental naive Bayes classifier to continuously learn positive transfer relationships between tasks. These methods belong to feedback-based task selection methods. They continuously capture the interaction between tasks by building an online learning model that dynamically captures task similarity. In addition, other researchers always adopt similarity-based task selection strategies. They gave a specific similarity measurement mechanism and used it to dynamically measure the relationship between tasks to select a suitable source task for the target task. For example, Tang \emph{et al.} proposed a group-based MFEA (GMFEA) to cluster similar tasks according to the distance between the best solutions \cite{ tangGroupbasedApproachImprove2018}. Wang \emph{et al.} designed a multi-task evolutionary algorithm based on anomaly detection (MTEA-AD) \cite{wangSolvingMultitaskOptimization2021}. Da \emph{et al.} captured the relationship between tasks by weight-index of probabilistic mixture model \cite{ daCurbingNegativeInfluences2019}. Zheng \emph{et al.} designed self-regulated evolutionary multitasking optimization (SREMTO), which replaces the factorial rank in MFEA by ability vector \cite{zhengSelfRegulatedEvolutionaryMultitask2020}.

Moreover, several studies have been devoted to suppressing negative transfer in terms of task transfer. These methods focus on generalizing useful knowledge and transferring them through specific transfer strategies. Ding \emph{et al.} proposed an EMTO framework to overcome the negative shift caused by the misalignment of the optimal solution \cite{ dingGeneralizedMultitaskingEvolutionary2019}. Bali \emph{et al.} established LDA-MFEA method by a linearized domain adaptation strategy \cite{baliLinearizedDomainAdaptation2017}. Zhou \emph{et al.} designed a new MFEA with adaptive knowledge transfer (MFEA-AKT). In addition to being extensively studied in MFO, task transfer has spawned another class of approaches to achieve EMTO, multi-population-based methods (MM). They assign a separate population to each task with a specific coding length and evolutionary operators. Knowledge transfer is achieved through explicit transfer strategies in MM. Feng \emph{et al.} proposed the first explicit transfer algorithm evolutionary multitasking via explicit autoencoding (EMEA) using autoencoders \cite{ fengEvolutionaryMultitaskingExplicit2019}. Then, Zhou \emph{et al.} expended EMEA by kernelized autoencoding and used a new method to solve multi-objective optimization problem \cite{ zhouLearnableEvolutionarySearch2021}. Wu \emph{et al.} applied orthogonal transfer to overcome the negative transfer caused by dimensional inconsistency \cite{wuOrthogonalTransferMultitask2022}. Inspired by symbiosis in biomes, Liaw \emph{et al.} proposed a new framework symbiosis in biocoenosis genetic algorithm (SBGA) to deal with the MaTO problem \cite{liawEvolutionaryManytaskingOptimization2019}.

These related works show that most existing EMTO methods focus on solving continuous optimization problems. Most of the existing similarity measures and transfer operators are designed for continuous coding. They cannot adapt to COP with discrete coding. Although there are some excellent approaches to multitasking COP, they are weak when dealing with many-task COP and cross-domain transfer. Feedback-based task selection mechanisms, redundant coding, and implicit transfer all lead to a large number of negative transfers, making transfer inefficient.

%---------------------------II.	B.	Related works---------------------------------
\subsection{Combinatorial optimization problem}
COP is one of the most important topics in the field of optimization. They have discrete decision variables and limited search space. Most of them belong to NP-hard problems that are difficult to solve in polynomial time. COP mainly includes two types of problems, permutation-based representation represented by path planning and discrete value-based representation represented by matching problem \cite{MachineLearningService2022}. This work involves two kinds of problems based on permutation-based representation: traveling salesman problem (TSP) \cite{osabaChapterTravelingSalesman2020}, capacitated vehicle routing problem (CVRP) \cite{toth2002vehicle} and two kinds of problems based on discrete value-based representation: quadratic assignment problem (QAP) \cite{lawlerQuadraticAssignmentProblem1963}, linear ordering problem (LOP) \cite{bertsimas1997introduction}. The details of these four problems are shown as follows.

TSP: Given a fully connected graph with $n$  cities, the $n\times n$ adjacency matrix of distances between cities is $D={{\{{{d}_{i,j}}\}}_{i,j\in \{1,...,n\}}}$ , where ${{d}_{i,j}}$ is the distance between city $i$ and city $j$. The objective function of TSP is to find a Hamiltonian cycle $S\text{=(}{{s}_{1}}\text{,}{{s}_{2}},...,{{s}_{n}}\text{)}$ with the smallest total distance, $S$ is a permutation of the city set. Mathematically, the objective function of TSP is:

\begin{equation}
	\label{eq2}
	f(S)=\sum\limits_{i=2}^{n}{{{d}_{{{s}_{i-1}},{{s}_{i}}}}}
\end{equation}

 CVRP: Given a fully connected graph with $n$  customers and a depot, the $(n+1)\times (n+1)$ adjacency matrix of distances between customers and the depot is $D={{\{{{d}_{i,j}}\}}_{i,j\in \{01,...,n\}}}$ , where ${{d}_{i,j}}$ is the distance between vertex $i$ and vertex $j$. The objective function of CVRP is to find a permutation of the customers set $S\text{=(}{{s}_{1}}\text{,}{{s}_{2}},...,{{s}_{n}}\text{)}$ with the shortest total distance. Where $S$ can be divided into multiple Hamiltonian cycles starting from the depot, and the demand for each cycle does not exceed the limit capacity.

 QAP: Given a $n\times n$ distance matrix $D={{\{{{d}_{i,j}}\}}_{i,j\in \{1,...,n\}}}$ and a flow matrix $C={{\{{{c}_{i,j}}\}}_{i,j\in \{1,...,n\}}}$, where ${{d}_{i,j}}$ is the distance between location $i$ and location $j$ and ${{c}_{i,j}}$ is the value of flow between facility $i$ and facility $j$ . The objective function of QAP is to assign $n$ facilities to $n$ locations with the smallest total cost. The solution is expressed as $S\text{=(}{{s}_{1}}\text{,}{{s}_{2}},...,{{s}_{n}}\text{)}$, where ${{s}_{k}}$ means assign the ${{s}_{k}}$-th facility to the $k$-th location.

 \begin{equation}
 	\label{eq3}
 	f(S)=\sum\limits_{i=1}^{n}{\sum\limits_{j=1}^{n}{{{c}_{{{s}_{i}},{{s}_{j}}}}{{d}_{{{s}_{i}},{{s}_{j}}}}}}
 \end{equation}

LOP: Given a $n\times n$ matrix $M={{\{{{m}_{i,j}}\}}_{i,j\in \{1,...,n\}}}$, The goal of LOP is to find a permutation of $S\text{=(}{{s}_{1}}\text{,}{{s}_{2}},...,{{s}_{n}}\text{)}$ of the rows and columns of $M$ which maximizes the sum of elements above the diagonal. The objective function is as follows:

 \begin{equation}
	\label{eq4}
	f(S)=\sum\limits_{i=1}^{n}{\sum\limits_{j=i}^{n}{{{m}_{{{s}_{i}},{{s}_{j}}}}}}
\end{equation}

In the global supply chain, it is assumed that there is a product production company with a complete logistics system. The MaTO scenario composed of these four problems is described in terms of the actual decision-making needs in its business.

To meet the commuting needs of employees, companies need to deploy multiple shuttles. The driving path of these vehicles is a Hamilton Circle containing the factories. Optimizing the routes of all shuttles can reduce commuting time and operating costs. It can be modeled as multitasking TSP. Logistics and transportation services in various regions are also important to the company's operating costs. The delivery plan of warehouse centers can be modeled as VRPs. Optimizing logistics solutions can allocate a reasonable number of vehicles to each warehouse and save transportation costs. Furthermore, the cost of dispatching goods between different storage centers is also part of the company's logistics and transportation costs. According to the transportation cost and actual transportation distance in each region, it is also important for the company to reasonably deploy the location of the warehouse centers in different regions. This problem can be modeled by multitasking QAP in practical optimization. Finally, the difference in the benefits of different product process arrangements in production cannot be ignored. In actual production, limited by the preparation of raw materials before the start of each process and the difference in machine efficiency, different process arrangements will have different benefits. It can be modeled as LOPs by optimizing the sequence of operations to maximize profit. These four optimization problems exist in the company's decision-making. Optimizing these issues concurrently will help decision-makers increase revenue and save costs.  Unfortunately, limited by the lack of actual business data, there is currently no real-world dataset to support the optimization requirements in this scenario.

Many related works have been proposed to address them. These methods can be roughly divided into three categories: exact algorithms, heuristics, and meta-heuristics \cite{MachineLearningService2022}. Exact algorithms can use mathematical methods to find the global optimal solution for COP, but the computational complexity is high, such as linear programming \cite{dantzigSolutionLargeScaleTravelingSalesman1954}, branch-and-bound method \cite{landAutomaticMethodSolving1960} and dynamic programming method \cite{bellmanDynamicProgrammingTreatment1962}. Moreover, heuristics are a class of iterative-based methods. They can iterate from a random solution to an approximate optimal solution in polynomial time, including variable neighborhood search (VNS) \cite{mladenovicVariableNeighborhoodSearch1997}, iterated local search (ILS) \cite{lourencoIteratedLocalSearch2019}, greedy heuristic (GH), greedy randomized adaptive search procedure (GRASP) \cite{feoGreedyRandomizedAdaptive1995}. There are also some local search operators designed for specific problems, such as k-opt \cite{croesMethodSolvingTravelingSalesman1958} and Lin–Kernighan heuristic (LKH) \cite{helsgaunEffectiveImplementationLin2000}. In addition, meta-heuristics are a class of population-based methods that do not rely on the gradient information of problems and are widely applied to solve various COPs. Representative algorithms include genetic algorithm (GA) \cite{lamboraGeneticAlgorithmLiterature2019}, particle swarm optimization (PSO) \cite{clercDiscreteParticleSwarm2004}, ant colony optimization (ACO) \cite{dorigoAntColonyOptimization2006} and artificial bee colony
(ABC) \cite{karabogaIDEABASEDHONEY2005}. Some researchers tried combining the meta-heuristic method with the heuristic method and designed the memetic algorithm (MA) \cite{moscatoEvolutionSearchOptimization1989}. In the family of MAs, many greedy-based heuristics are used to initialize individuals in the population, and single-solution-based heuristics are often applied as local search. By combining global and local search, MAs can find solutions with better quality. Inspired by this, heuristics' low computational complexity and local search capability provide a natural idea for EMTO to suppress negative transfer with a small computational cost in COP.

\section{Proposed Method}
This section introduces the details of each part of MTEA-AST. The complexity analysis of key operators in MTEA-AST is also presented in this section.

\subsection{Dimension Unification Strategy}
The dimension of the search space for different tasks may not be uniform in EMTO, which leads to heterogeneous characteristics of different individuals. Therefore, unifying the dimensions of various tasks is a significant issue. Most existing methods for cross-domain COPs use the unified search space strategy and introduce random coding to individuals in low-dimensional tasks. It also introduces meaningless noise information that leads to inefficient transfer.

In this work, the idea of a greedy heuristic is applied to unify the dimension and introduce prior information to replace the noise in random coding. Given a source task ${{T}_{s}}$ and a target task ${{T}_{t}}$, their dimensions are ${{D}_{s}}$, ${{D}_{t}}$, the corresponding solutions are permutations of sets $S_{s}=\left\{1,2, \ldots, D_{s}\right\}$ and  $S_{t} =\left\{1,2, \ldots, D_{t}\right\}$. $x$ is a solution for the source task. Assuming ${{D}_{s}}<{{D}_{t}}$, $\varphi=S_{t}-S_{s}$, all elements in $\varphi$ are inserted into the current solution  in turn according to the principle of minimizing cost increase. There are an element ${{\varphi }_{i}}\in \varphi $, ${{f}_{t}}(.)$ is the objective function of the target task, ${{x}^{k}}$  is the new solution obtained by inserting ${{\varphi }_{i}}$ into the $k$-th position of the current solution $x$. And the principle for the insertion position of ${{\varphi }_{i}}$ is shown in Eq. (\ref{eq5}).

 \begin{equation}
	\label{eq5}
	k=\arg {{\min }_{k}}{{f}_{t}}({{x}^{k}})-{{f}_{t}}(x)
\end{equation}

\begin{figure}[htp]
    \begin{center}
        \centerline{\includegraphics[width=0.79\linewidth]{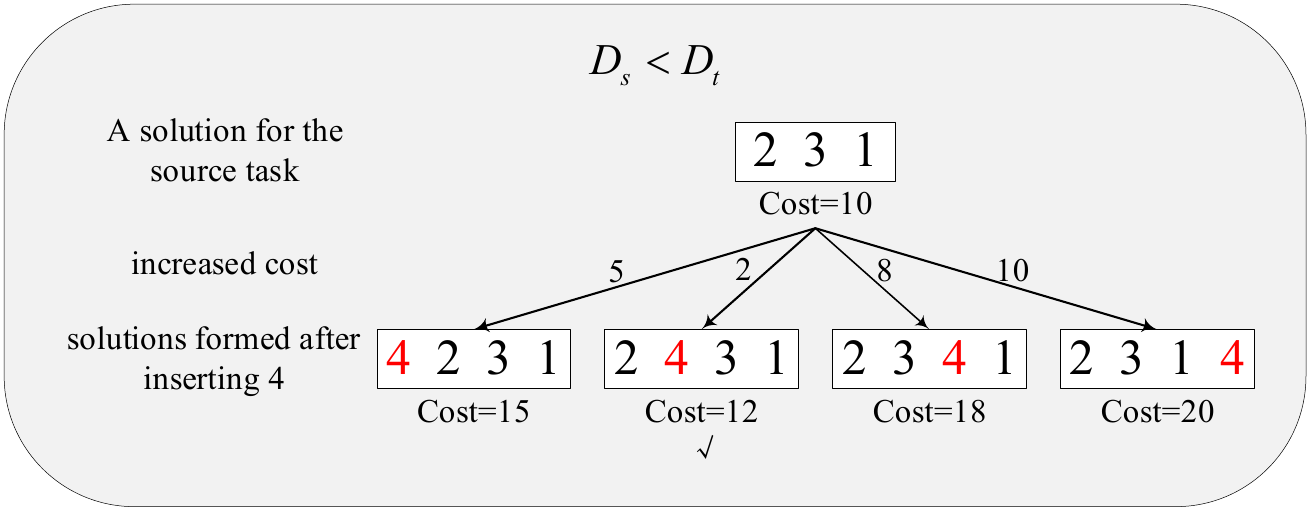}}
        \caption{ Examples of dimension unification strategies.\label{fig6}}
    \end{center}
\end{figure}

As is shown in Fig. \ref{fig6}, assuming the dimension of the source task ${{D}_{s}}=3$ and the dimension of the target task ${{D}_{t}}=5$. A solution [2,3,1] responsible for the source task needs to be mapped into the target task search space. The set $\varphi$ of elements to be inserted is [4,5]. Firstly, element 4 is inserted into different positions of the solution [2,3,1] to form different candidate solutions. And then, [2, 4, 3, 1] with the smallest cost is selected as the initial solution for the next insertion of element 5. After that, insert element 5 according to the same rules to complete this mapping process. In another case, ${{D}_{s}}>{{D}_{t}}$, only keep elements in $x$ that are less than ${{D}_{t}}$. 

% The details of the dimension unification strategy are shown in Algorithm \ref{alg1}:

% %算法1
% \begin{algorithm}
% 	\renewcommand{\algorithmicrequire}{\textbf{Input:}}
% 	\renewcommand{\algorithmicensure}{\textbf{Output:}}
% 	\caption{Dimension unification strategy \label{alg1}}
% 	\begin{algorithmic}[1]%一行一个标行号
% 		\REQUIRE $x$ (individual in the source population), $ f_{t}(.)$ (the objective function of the target task), ${{D}_{s}}$ ( dimension of the source task), ${{D}_{t}}$ ( dimension of the target task)
% 		\ENSURE $x$ (the mapped solution for the target task)
% 		\IF{${{D}_{s}}<{{D}_{t}}$}
% 		\STATE $\varphi=\left\{ D_{s}+1, D_{s}+2, \ldots, D_{t}\right\}$;
% 		\WHILE{$\varphi \ne \phi$}{\item
% 		Insert the element ${{\varphi }_{i}}\in \varphi$ into $x$ in sequence according to Eq (\ref{eq5}) and remove ${{\varphi }_{i}}$ from $\varphi$;}
%         \ENDWHILE
% 		\ELSE
% 		\STATE Delete all elements that are smaller than ${{D}_{t}}$ in $x$;
% 		\ENDIF
% 	\end{algorithmic}
% \end{algorithm}

%---------------------------III.	B.	LSN---------------------------------
\subsection{Adaptive Task Selection Strategy}
The dimension unification strategy enables cross-task individual evaluation. However, a large number of tasks are simultaneously optimized in the MaTO scenario. The negative transfer caused by blind knowledge transfer will greatly weaken the effect of positive transfer. To address this issue, most of the existing adaptive task selection strategies adopt feedback-based methods, which need to collect interactive information between tasks and update the model online. Therefore, there is still a blind transfer in the early stage of transfer. Similarity-based methods are another way to deal with MaTO. With the help of prior knowledge, they can achieve significant effects in the early stage of optimization. According to \cite{ zhouStudySimilarityMeasure2018}, the similarities between continuous optimization tasks are mainly reflected in three aspects, i.e., the distance between the best solutions, the fitness rank correlation, and the fitness landscape analysis. COP is difficult to obtain the landscape features, and fitness rank requires repeat cross-task individual evaluation. Therefore, the distance of the optimal solution between tasks is an excellent principle to measure the similarity between tasks.

Our proposed adaptive task selection strategy mainly consists of a cross-task individual representation and a similarity measure mechanism. The optimization problem cannot obtain information on the global optimal solution before optimization. Unlike the probabilistic crossover in implicit transfer strategies, the MM framework usually exchanges information with a specific frequency. Individuals in different populations have enough time to develop suitable characteristics for the corresponding tasks. In COP, some graph structures of the current best solutions in the population are the same as the optimal solutions, and the distance between the local optimal solution and the global optimal solution will be smaller with the optimization process. so we use the Hamming distance between the best solutions in each population to replace the distance of the global optimal solution. In addition, this work studies the permutation-based problem and the discrete value-based problem. The rotation invariance of the former solution diversifies the representation of individuals. An adjacency matrix-based individual representation is used to solve this issue.

\begin{figure*}[htp]
	\centering
	\includegraphics[width=0.97\textwidth]{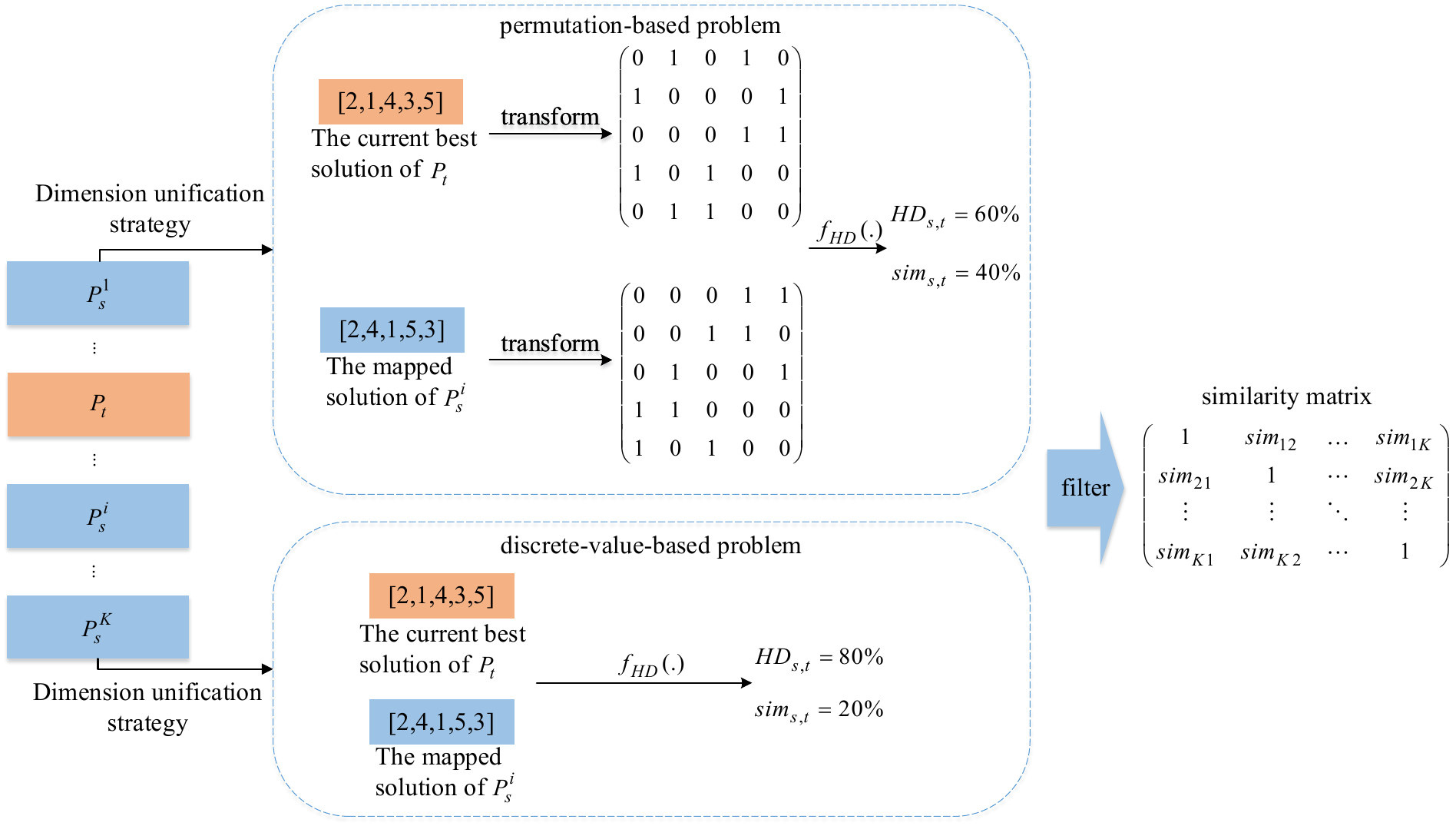}
	\caption{The overflow of the adaptive task selection strategy.\label{fig1}}
\end{figure*}

The overflow of the adaptive task selection strategy is shown in Fig. \ref{fig1}. First, the target task ${{T}_{t}}$ is selected, and the current best solution in each source task is mapped through the dimensional unification strategy. Second, the mapped solution transforms the representation of individuals according to the target task type. Specifically, in permutation-based problems, individuals are transformed into binary adjacency matrices to overcome rotation invariance, while solutions to discrete-valued-based problems remain unchanged. Next, the transformed individuals are used to calculate the Hamming distance between the current best solutions according to Eq. (\ref{eq6}). 

\begin{equation}
\label{eq6}
% \resizebox{0.9\hsize}{!}{
H{{D}_{i,j=}}\left\{ \begin{aligned}
	& \frac{{{f}_{HD}}({{s}_{i}},{{s}_{j}})}{2\times {{D}_{t}}}\text{ }(\text{permutation-based problem}) \\ 
	& \frac{{{f}_{HD}}({{s}_{i}},{{s}_{j}})}{{{D}_{t}}}\text{ }(\text{discrete-valued-based problem}) \\ 
\end{aligned} \right.
% }
\end{equation}

Where ${{HD}_{i,j}}$ is the Hamming distance between the target task and the source task, and ${{s}_{i}}$ and ${{s}_{j}}$ are the transformed current best solutions of the two tasks. ${{f}_{HD}}(.)$ is the formula for calculating the Hamming distance. ${{D}_{t}}$ is the dimension of the target task. It is worth noting that the transformed adjacency matrix is symmetric in permutation-based problems. Therefore, it needs to be halved after calculating the Hamming distance of the two matrices. The similarity between these two tasks is calculated by Eq. (\ref{eq7}):

\begin{equation}
\label{eq7}
	{{sim}_{i,j}}=1-H{{D}_{i,j}}
\end{equation}

So far, a similarity matrix reflecting the relationship between all tasks has been obtained. According to \cite{osabaMultifactorialCellularGenetic2020}, a similarity of 10$\%$ is sufficient to maintain minimal positive transfer effects. Screen out task pairs whose similarity is less than 10$\%$ in the similarity matrix. It ensures the transfer effect while filtering out a large number of low-similarity tasks and avoids the computational burden caused by repeated cross-task individual evaluations. Finally, the transfer strength is calculated by the similarity between the target task and each source task, which is applied to selecting candidate solutions for seeds. Its calculation formula is shown in Eq. (\ref{eq8}).

\begin{equation}
	\label{eq8}
	{{p}_{i,j}}=\varepsilon \times \frac{si{{m}_{i,j}}}{\sum\limits_{j=1j\ne i}^{K}{si{{m}_{i,j}}}}
\end{equation}

Where ${{p}_{i,j}}$ is the transfer strength between the $i$-th task and the $j$-th task, i.e., the number of seed candidates. $\varepsilon $ is a pre-specified value that controls the number of candidate solutions picked from each source task for a target task. $si{{m}_{i,j}}$ is the similarity between the $i$-th task and the $j$-th task.

% 图1

%---------------------------III.	C.	proposed method---------------------------------
\subsection{Task Transfer Strategy}
To suppress negative transfer caused by cross-domain transfer between heterogeneous problems and adapt to MaTO, how to transfer useful knowledge is also important in EMTO. Implicit transfer strategies usually perform probabilistic crossover of individuals responsible for different tasks to share useful knowledge. These knowledge transfer operators lead to serious negative transfer in the case of knowledge differences between individuals. Knowledge transfer is usually achieved by inserting excellent individuals in explicit transfer mechanisms. The migrated individuals can change the target population's distribution and guide the offspring's generation through their outstanding performance. Existing MMs often use the k-best individual selection strategy, which is effective when the similarity between tasks is high. But in the case of low similarity, especially in the early stage of optimization, the chromosomes that perform well in the source task may not be the focus of the target task. Instead, there may be some useful chromosomes for the target task in some suboptimal solutions.

 The definition of ability vector is given as follows:
 
$Ability\text{ }Vector$: ${{v}_{i}}=\{v_{i}^{j}\}_{j=1}^{K}$ is the ability vector of the $i$-th individual. $v_{i}^{j}\text{=}{{f}_{m}}(r_{i}^{j})$, where ${{f}_{m}}(.)$ is a piecewise monotonically decreasing function and $r_{i}^{j}$ is the factorial rank of the $i$-th individual in the $j$-th task. 

The ability vector is obtained through the cross-task individual evaluation of the source and target tasks. It assigns the attention of the transfer strategy to some better individuals and retains the probability of other solutions being selected. This idea is similar to the fitness rank correlation-based similarity. However, ability vectors require cross-task evaluation of individuals in two populations, which is computationally expensive. In this work, we propose the concept of ability fitness. It is the scalar mean of the individual's ability vector in the target and source tasks, reflecting the individual's adaptive capability to the two tasks. Some poor solutions in the source task population cannot obtain high ability fitness and do not require cross-task individual evaluation, thus computational resources are saved. Given a target task ${{T}_{t}}$ and a source task ${{T}_{s}}$, the corresponding populations are ${{P}_{t}}$ and ${{P}_{s}}$. The similarity between ${{T}_{t}}$ and each source task obtained according to the adaptive task selection strategy is $si{{m}_{t}}$, and the transfer strength between ${{T}_{t}}$ and ${{T}_{s}}$ is ${{p}_{s,t}}$, that is, ${{p}_{s,t}}$ outstanding individuals need to be selected from ${{P}_{s}}$ and added to the seed candidate set ${{\psi }_{t}}$ of ${{T}_{t}}$. According to the seed selection mechanism based on ability fitness, the ability vectors of the first $2\times {{p}_{s,t}}$ individuals in ${{P}_{s}}$ are first calculated. Then, the ability fitness is calculated based on the ability vector, the individuals are sorted in descending order according to the ability fitness, and the first ${{p}_{s,t}}$ individuals are selected to join ${{\psi }_{t}}$. After selecting from the various source task populations, ${{\psi }_{t}}$ will have $\varepsilon $ seed candidates. Next, $\lambda $ seeds are generated from the seed candidate set ${{\psi }_{t}}$ according to the value of ability fitness.

It should be noted that selecting all individuals in the candidate set according to their ability fitness is conducive to selecting excellent individuals from multiple source tasks, so that the target task can receive information from different source tasks and maintain the diversity of the population. Furthermore, a new seed growth strategy is designed to further optimize these seeds by adopting matching operators from a variety of local search operators according to the target task. Seed growth operators are different from ordinary local search operators embedded within evolutionary operators. Specifically, the local search operators are limited to the maximum iteration to reduce the computational burden. Since the number of seeds is far less than the population size, the seed growth operator can use local search operators with relatively high computational complexity to achieve a better search effect. This strategy can guarantee the adaptability of seeds to the target task population and provide good guidance for the target task in the early stage of optimization. Finally, seeds are inserted into the ${{P}_{t}}$ to replace the original individuals with the closest Hamming distance. The details of the task migration strategy are shown in Algorithm \ref{alg2}.

%算法2
\begin{algorithm}[htp]
	\renewcommand{\algorithmicrequire}{\textbf{Input:}}
	\renewcommand{\algorithmicensure}{\textbf{Output:}}
	\caption{Task Transfer Strategy \label{alg2}}
	\begin{algorithmic}[1]%一行一个标行号
		\REQUIRE ${{p}_{s,t}}$ (the transfer strength  between the target task and source tasks), ${{f}_{s}}(.)$ (the objective function of the source task), ${{P}_{t}}$ (the population of the target task), ${{P}_{s}}$ (the population of the source task), $\varepsilon $(the number of seed candidates), $\lambda $ (the number of seeds)
		\ENSURE ${{P}_{t}}$ (the population of the target task after transfer)
		\WHILE{the number of seed candidates in ${{\psi }_{t}}$ is less than $\varepsilon $}{
		\STATE given a source task ${{T}_{s}}$;
		\IF{${{p}_{s,t}}\ne 0$}
		\STATE Sort ${{P}_{s}}$ in ascending order according to ${{f}_{s}}(.)$; 
		\STATE Calculate the ability vector of the first $2\times {{p}_{s,t}}$ individuals;
		\STATE Calculate the ability fitness of the first $2\times {{p}_{s,t}}$ individuals;
		\STATE Select top ${{p}_{s,t}}$ individuals to join ${{\psi }_{t}}$ according to their ability fitness;
		\ENDIF}
		\ENDWHILE 
		\STATE Select top $\lambda $ seeds in ${{\psi }_{t}}$ according to their ability fitness;
		\STATE All seeds perform a local search through the seed growth strategy;\\
		\STATE All seeds are inserted into ${{P}_{t}}$ to replace the original individual with the closest Hamming distance;
	\end{algorithmic}
\end{algorithm}

%---------------------------III.	D.	Discussion of computational efficiency---------------------------------
\subsection{Framework of MTEA-AST}
\begin{figure*}[ht]
	\centering
	\includegraphics[width=0.97\textwidth]{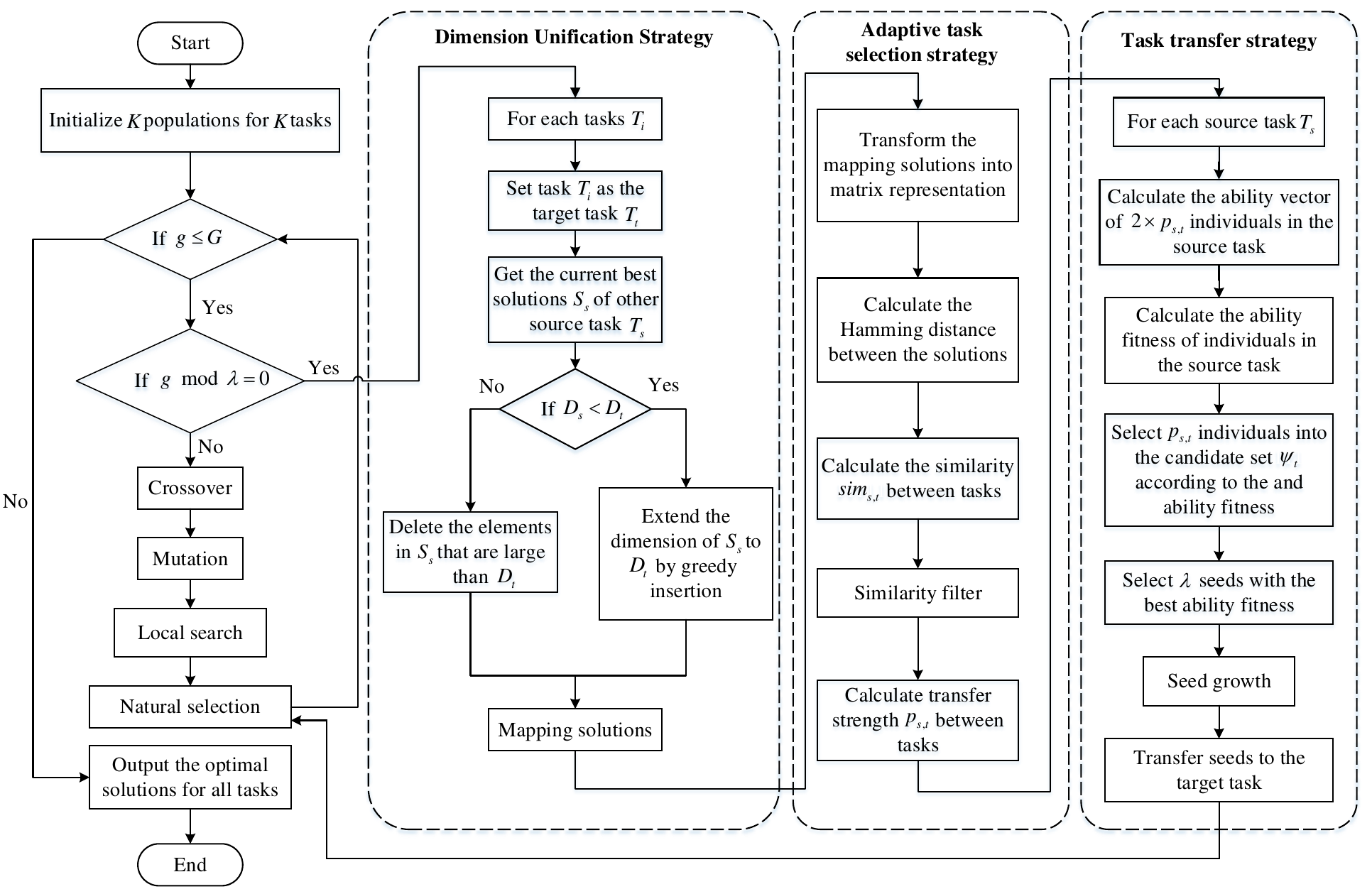}
	\caption{Framework details of MTEA-AST.\label{framework}}
\end{figure*}

The framework of MTEA-AST is given in Algorithm \ref{alg3} and Fig. \ref{framework}. Given $K$ COPs, the size of each population responsible for each task is $N$. Furthermore, the number of seed candidates is $\varepsilon $, the number of seeds is $\lambda $, and the transfer frequency is $\alpha $. Firstly, $K$ separate populations are initialized, and multiple evolutionary solvers are used to optimize each task. When the algorithm iterates for $\alpha $ generations, the current optimal individuals in $K$ populations are picked out. Then, each task is set as the target task ${{T}_{t}}$ in turn, and its dimension is ${{D}_{t}}$. All current best solutions from other tasks are mapped to ${{D}_{t}}$ through the proposed dimension unification strategy. Next, the current best solution of the target task and other mapped solutions are used to calculate the similarity $si{{m}_{t}}$ between ${{T}_{t}}$ and other online tasks through the adaptive task selection strategy. Based on the similarity, the transfer strength ${{p}_{s,t}}$ is obtained, i.e., the number of the seed candidates selected from each source task population ${{P}_{s}}$. Next, according to the transfer strategy, the ability fitness of the top $2\times {{p}_{s,t}}$ individuals is calculated in each ${{P}_{s}}$, and the top ${{p}_{s,t}}$ individuals are selected to join the seed candidate set ${{\psi }_{t}}$ of the target task. Furthermore, selecting $\lambda $ seeds with the most excellent ability fitness in ${{\psi }_{t}}$, and the seed growth strategy is applied to further optimize the seeds to adapt to ${{P}_{t}}$. Finally, all seeds are inserted into ${{P}_{t}}$ to replace the same number of original individuals with the closest Hamming distance.

\begin{algorithm}[ht]
	\renewcommand{\algorithmicrequire}{\textbf{Input:}}
	\renewcommand{\algorithmicensure}{\textbf{Output:}}
	\caption{MTEA-AST \label{alg3}}
	\begin{algorithmic}[1]%一行一个标行号
		\REQUIRE $\{{{T}_{k}}={{f}_{k}}(.),\text{ }1\le k\le K\}$ ($K$ tasks and objective functions), $N$ (the population size of each task), $G$ (the maximum generation), $\varepsilon $ (the number of seed candidates), $\lambda $ (the number of seeds), $\alpha $( transfer frequency)
		\ENSURE ${{X}^{\text{*}}}=\{x_{k}^{*},\text{ }1\le k\le K\}$ (the best solutions of all tasks)
		\STATE $\{{{P}_{k}},\text{ }1\le k\le K\}\leftarrow \text{Initialize }K\text{ populations}$;
		\STATE $g\leftarrow 0$ 
		\WHILE{$g \le G$}{
			\IF{$g \ mod \ \lambda \neq 0$}
                \FOR{$i=1\text{ to }K$}
			\STATE${{O}_{i}}\leftarrow$ Perform crossover mutation and local search on ${{P}_{i}}$;
			\STATE Evaluate individuals’ fitness in ${{O}_{i}}\cup {{P}_{i}}$ according to ${{f}_{i}}(.)$;
			\STATE Select top $N$ individual from ${{O}_{k}}\cup {{P}_{k}}$ based on their fitness value;
                \ENDFOR
			\ELSE 
			\FOR{$i=1\text{ to }K$}
			\STATE Target task ${{T}_{t}}\leftarrow {{T}_{i}}$;
                \STATE $\{{{s}_{k}},k \neq i\}\leftarrow $the current best solutions from other populations $\{{P}_{k}, k \neq i\}$;
			\STATE Map ${{s}_{k}}$ to ${{D}_{k}}$ dimension by the unified dimension strategy; %(see Algorithm \ref{alg1})
			\STATE Calculate the similarity $si{{m}_{s,t}}$ and transfer strength ${{p}_{s,t}}$ between ${{T}_{t}}$ and other source tasks by the adaptive task selection strategy;\\
			\STATE Select $\lambda $ seeds from each source population ${{P}_{s}}$ and transfer them into target population ${{P}_{t}}$ after performing seed growth operator (see Algorithm \ref{alg2})
			\ENDFOR
	    \ENDIF}
            \STATE $g\leftarrow g+1$ 
		\ENDWHILE
	\end{algorithmic}
\end{algorithm}

For knowledge transfer in one generation, the computational cost in the transfer process mainly consists of two parts: the dimension unification strategy and the task transfer strategy. The time complexity of the dimension unification strategy in the adaptive task selection strategy is $O({{K}^{2}}{{D}^{2}})$, where $K$ and $D$ are the number of tasks and the number of dimensions in each task. In the task transfer strategy, the computational cost involves the seed candidate selection mechanism and seed growth  operator. The time complexity of the seed candidate selection mechanism is $O(\varepsilon K{{D}^{2}})$, where $\varepsilon $, $K$ and $D$ are the number of seed candidates, the number of tasks and the dimensions of each task. Moreover, the time complexity of the seed growth operator is $O(\lambda K{{D}^{3}})$, where $\lambda $, $K$ and $D$ are the number of seeds, the number of tasks and the dimensions of each task.
%-------------------------------------------------------------IV. Experimental Design------------------------------------------------
\section{Experimental Results and Analysis}
To investigate the performance of MTEA-AST, experiments are designed in this work to compare the proposed method with several state-of-the-art EMTO methods on same-domain MaTO benchmarks and cross-domain MaTO benchmarks. Firstly, the comparison results of solution quality and search efficiency between MTEA-AST and other methods in same-domain MaTO scenarios are analyzed (\ref{4.b}). And then, the comparison results in cross-domain MaTO scenarios are given (\ref{4.c}). Next, the feasibility of the proposed adaptive task selection strategy is discussed (\ref{4.d}). Moreover, the results of ablation experiments are presented to verify the effectiveness of the dimension unification strategy and task transfer strategy (\ref{4.e}). Finally, the analysis on the relationship between transfer effect and task similarity is given (\ref{4.f}).

%---------------------------IV.	A.	Peer Competitors---------------------------------
\subsection{Experimental Settings}
In this work, the many-task dataset proposed by \cite{osabaATMFCGAAdaptiveTransferguided2021} is adopted to compare the performance of MTEA-AST with five other methods. This dataset involves 4 different types of COPs. It is noted that LOP is a maximization problem, the fitness function of the optimizer corresponding to this problem is set to the opposite of its objective function to convert it into a minimization problem. The experimentation consists of 11 test cases, which can be divided into 4 same-domain many-task benchmarks and 7 cross-domain many-task benchmarks. The details of the dataset are shown in  Table \ref{table1}.

\begin{table}[htbp]
	\renewcommand{\arraystretch}{1.2}
	\caption{An overview of the dataset used for the experiments.\label{table1}}
	\centering
		\resizebox{0.93\textwidth}{!}{ % 设置表格宽度
	\begin{tabular}{ccc}
		\hline
		Type                               & Test Case & Instances involved                                         \\ \hline
		\multirow{4}{*}{Same-domain MaTO}  & TSP       & kroA100, kroA150, kroA200, kroB150, kroC100                \\
		& CVRP      & P-n50-k7, P-n50-k8, P-n55-k7, P-n55-k15, P-n60-k10         \\
		& QAP       & Nug25, Nug30, Kra30a, Kra30a, Kra32a                       \\
		& LOP       & N-t59d11xx, N-t59f11xx, N-t59i11xx, N-t65f11xx, N-t70f11xx \\
		\multirow{7}{*}{Cross-domain MaTO} & TSP\_CVRP & TSP$\cup $CVRP                                          \\
		& TSP\_QAP  & TSP$\cup $QAP                                           \\
		& TSP\_LOP  & TSP$\cup $LOP                                           \\
		& CVRP\_QAP & CVRP$\cup $QAP                                          \\
		& CVRP\_LOP & CVRP$\cup $LOP                                          \\
		& QAP\_LOP  & QAP$\cup $LOP                                           \\
		& ALL       & TSP$\cup $CVRP$\cup $QAP$\cup $LOP                \\ \hline
	\end{tabular}}
\end{table}

 Four state-of-the-art EMTO methods for cross-domain many-task COP and a traditional single-task optimization (STO) method are applied as experimental baselines: MFEA\cite{yuanEvolutionaryMultitaskingPermutationbased2016}, DMFEA-II\cite{osabaDMFEAIIAdaptiveMultifactorial2020}, MFCGA\cite{osabaMultifactorialCellularGenetic2020}, AT-MFCGA\cite{osabaATMFCGAAdaptiveTransferguided2021} and STO method. MFEA is the earliest EMTO framework and was first applied to COPs. DMFEA-II improves MFEA by a designed negative feedback mechanism. MFCGA is used to solve the multi-task traveling salesman problem. On this basis, AT-MFCGA realizes the adaptive knowledge transfer by dynamically adjusting the neighbor structure and is applied to the cross-domain transfer involving four COPs. In addition, the Hybrid Genetic Algorithm \cite{yuanEvolutionaryMultitaskingPermutationbased2016,fengExplicitEvolutionaryMultitasking2021} was set as STO to verify the feasibility of transfer operator in MTEA-AST. In order to ensure the fairness of comparison, the evolution operators and local search operators used by all EMTO frameworks are consistent with STO. They all use order crossover and swap mutation. In terms of local search operators, 2-opt is adopted for TSP and CVRP \cite{1958TSP, VRP2019}, the swap operator is used in QAP \cite{BLS2013}, and the insertion operator is applied in LOP \cite{LOP2015}. All algorithms run with Python 3.8 and execute on an Ubuntu 18.04.2 PC with an Intel Core i7-8700K CPU, 3.7GHz, and 64GB RAM. All parameter settings are given in Table \ref{table5}.

a)	The population size of each task in MTEA-ST and single-task GA is set as 30, and the population size in MFEA, DMFEA-II, MFCGA, AT-MFCGA is 30$\times$k, where k is the number of tasks.

b)	The maximum generation for all algorithms is 300

c)	The number of independent runs for all methods is 20.

d)	For parameters related to knowledge transfer, the number of seed candidates $\varepsilon $ for MTEA-AST is 10, and the number of seeds $\lambda $ is 3. In order to maximize the performance of the transfer strategy, the relevant parameters of other EMTO methods are set according to the original text.

\begin{table*}[h]
\centering
\renewcommand{\arraystretch}{1.3}
	\caption{Parameter settings of all algorithms. \label{table5}}
    \scriptsize
	% \resizebox{\textwidth}{!}{ % 设置表格宽度
    \setlength{\tabcolsep}{1mm}{
\begin{tabular}{ccccccc}
 \hline
Parameter          & MTEA-AST   & STO  & AT-MFCGA  & MFCGA  & DMFEA-II & MFEA \\  \hline
Population size & 30*k             & 30*k  & 30*k             & 30*k       & 30*k          & 30*k  \\
Generation        & 300               & 300    & 300              & 300        & 300            & 300  \\
Crossover          & order crossover  & order crossover  & order crossover & order crossover & order crossover & order crossover \\
Mutation          & swap mutation    & swap mutation    & swap mutation    & swap mutation   & swap mutation   & swap mutation\\
Local search     & \begin{tabular}[c]{@{}c@{}}\{2-opt,\\ swap,\\ insertion\}\end{tabular}  & \begin{tabular}[c]{@{}c@{}}\{2-opt,\\ swap,\\ insertion\}\end{tabular}  & \begin{tabular}[c]{@{}c@{}}\{2-opt,\\ swap,\\ insertion\}\end{tabular}  & \begin{tabular}[c]{@{}c@{}}\{2-opt,\\ swap,\\ insertion\}\end{tabular}  & \begin{tabular}[c]{@{}c@{}}\{2-opt,\\ swap,\\ insertion\}\end{tabular}  & \begin{tabular}[c]{@{}c@{}}\{2-opt,\\ swap,\\ insertion\}\end{tabular} \\
Transfer frequency   & 10       & -   & -   & -   & -  & -       \\
${Num}_{Seed\_Candidate}$    & $\varepsilon=10 $          & -         & -      & -    & -    & -          \\
${Num}_{seed}$                       & $\lambda=3 $               & -          & -      & -     & -     & -          \\ 
Neighbor   & -   & -   & Moore  & Moore  & -   & -   \\
Adaptive frequency  & -   & -    & 10  & -   & -   & -  \\
${P}_{same\_task}$     & -   & -    & 0.5 & -   & -   & -  \\
Initial values of RMP & -   & -    & -     & -      & 0.95 & - \\
${\Delta}_{inc} $ / ${\Delta}_{dec} $  & -   & -   & -   & -   & 0.99/0.99  & - \\
RMP   & -   & -   & -   & -   & -  & 0.9 \\ 
\hline           
\end{tabular}}
\end{table*}

%---------------------------IV.	B.	Benchmark Datasets---------------------------------
\subsection{Results and discussion in same-domain MaTO problem} \label{4.b}
This section shows the comparison results of solution quality and search efficiency between MTEA-AST and other methods in same domain MaTO problems.
\subsubsection{Solution quality}

Table \ref{table2} shows the solution quality of MTEA-AST and five other compared methods on four same-domain many-task benchmarks. The results include the average and standard deviation of the objective function values for all algorithms. Symbols "$-$", "$\approx$" and "$+$" represent the corresponding methods are worse, similar, and better than MTEA-AST on the Wilcoxon rank-sum test with confidence level 95$\%$, respectively. In addition, the best performance results for each problem are identified in bold font.

\begin{table*}[htbp]
	\renewcommand{\arraystretch}{1.2}
	\caption{Comparative experimental results of all methods on benchmarks of same-domain MaTO problems in terms of objective value (average ± standard deviation).\label{table2}}
	\centering
	\resizebox{0.9\textwidth}{!}{% 设置表格宽度
		\begin{tabular}{cccccccc}
			\hline
			Test case             & Instance & MTEA-AST                     & STO                           & AT-MFCGA               & MFCGA                   & DMFEA-II                 & MFEA                   \\ \hline
			\multirow{5}{*}{TSP}  & T1       & \textbf{21930.7±453.41}      & 22588.16±520.13(-)            & 23105.36±473.08(-)     & 22974.35±778.72(-)    & 22990.89±579.85(-)          & 26613.14±1373.24(-)     \\
                      & T2       & \textbf{27901.72±296.43}     & 29386.05±907.11(-)            & 36484.72±1073.75(-)    & 36734.05±2077.95(-)   & 35682.61±1088.46(-)         & 45845.5±1856.12(-)      \\
                      & T3       & \textbf{31446.43±578.43}     & 42227.22±2718.53(-)           & 51366.59±1232.41(-)    & 51008.62±2245.28(-)   & 50179.49±1379.25(-)         & 63071.51±1674.14(-)     \\
                      & T4       & \textbf{27215.68±336.55}     & 28880.34±536.45(-)            & 35776.4±1400.34(-)     & 35514.38±2348.94(-)   & 34652.86±924.82(-)          & 44506.9±1228.3(-)       \\
                      & T5       & \textbf{21304.02±456.16}     & 22332.11±704.82(-)            & 22569.21±648.59(-)     & 22842.96±501.1(-)     & 22490.58±603.25(-)          & 26064.39±1350.62(-)     \\ \hline
\multirow{5}{*}{CVRP} & T1       & \textbf{623.91±18.58}        & 679.96±26.17(-)               & 769.23±19.74(-)        & 775.02±25.25(-)       & 736.41±27.4(-)              & 756.74±31.99(-)         \\
                      & T2       & \textbf{707.29±12.8}         & 754.23±16.52(-)               & 827.7±21.57(-)         & 835.74±26.6(-)        & 799.06±33.72(-)             & 820.61±31.51(-)         \\
                      & T3       & \textbf{652.02±19.41}        & 727.59±36.92(-)               & 847.17±23.05(-)        & 858.3±26.4(-)         & 793.79±30.52(-)             & 837.26±44.92(-)         \\
                      & T4       & \textbf{1025.73±17.89}       & 1084.21±27.53(-)              & 1163.42±16.86(-)       & 1154.97±20.2(-)       & 1123.83±27.19(-)            & 1150.25±17.99(-)        \\
                      & T5       & \textbf{840.8±15.98}         & 945.26±31.41(-)               & 1111.58±34.8(-)        & 1102.23±30.43(-)      & 1057.99±27.48(-)            & 1109.04±37.93(-)        \\ \hline
\multirow{5}{*}{QAP}  & T1       & 3860.8±62.17                 & 3859.9±48.45($\approx$)               & 3899.8±25.95(-)        & 3887.9±34.79(-)       & \textbf{3847.9±50.82($\approx$)}    & 3992.4±49.41(-)         \\
                      & T2       & 6337.1±83.45                 & 6358.0±99.63($\approx$)               & 6439.5±45.56(-)        & 6436.1±39.78(-)       & \textbf{6332.7±62.26($\approx$)}    & 6583.1±79.87(-)         \\
                      & T3       & 94816.0±1299.16              & 95394.0±1751.96($\approx$)            & 96036.5±1208.81(-)     & 96084.0±1018.98(-)    & \textbf{94233.0±1353.54($\approx$)} & 99168.0±1576.09(-)      \\
                      & T4       & 96038.0±1216.25              & 96573.0±1479.05($\approx$)            & 96558.5±861.56($\approx$)      & 96737.0±1552.29(-)    & \textbf{95630.5±1268.85($\approx$)} & 100337.5±1224.82(-)     \\
                      & T5       & 94530.0±1831.59              & \textbf{93825.0±1994.79($\approx$)}   & 96245.5±1026.48(-)     & 95841.5±696.48(-)     & 94228.0±1509.52($\approx$)          & 98895.0±1623.39(-)      \\ \hline
\multirow{5}{*}{LOP}  & T1       & \textbf{-146783.7±503.09}    & -146735.1±371.61($\approx$)           & -146935.55±352.55($\approx$)   & -146671.55±372.13($\approx$)  & -146782.2±313.26($\approx$)         & -144750.55±770.28(-)    \\
                      & T2       & \textbf{-122436.6±67.18}     & -122398.7±113.23($\approx$)           & -122253.6±137.43(-)    & -122157.95±185.36(-)  & -122428.1±110.17($\approx$)         & -121388.95±491.23(-)    \\
                      & T3       & \textbf{-8259188.35±3518.84} & -8243568.5±22153.16(-)        & -8244606.45±8005.64(-) & -8242313.4±13789.5(-) & -8251053.2±19306.77(-)      & -8198258.05±28674.95(-) \\
                      & T4       & \textbf{-216755.25±284.7}    & -216736.9±181.39($\approx$)           & -215834.75±572.59(-)   & -215823.6±657.79(-)   & -216430.55±344.98(-)        & -213755.35±1381.7(-)    \\
                      & T5       & -359312.25±674.67            & \textbf{-359327.55±537.79($\approx$)} & -357131.2±613.3(-)     & -357173.65±690.86(-)  & -358681.55±787.87(-)        & -353515.4±1765.39(-)    \\ \hline
			\multicolumn{8}{l}{“$-$”, “$\approx$” and “$+$”   indicate that the corresponding compared method is significantly worse,   similar, and better than MTEA-AST, respectively.}                                   
		\end{tabular}}
\end{table*}

According to Table \ref{table2}, MTEA-AST outperforms other EMTO methods on most same-domain benchmarks. Specifically, compared to AT-MFCGA, our proposal performs better on 18 and ties 2 out of 20 tasks in terms of the averaged objective value. Compared to DMFEA-II, it performers better on 13 and ties 7 in 20 tasks. Furthermore, MTEA-AST outperformers than MFCGA and MFEA in almost all 20 tasks. It means that our designed transfer operators are more efficient than other EMTO methods. Moreover, STO also shows advantages compared to other EMTO methods except MTEA-AST, especially in high dimensional problems (TSP, CVRP).  This is mainly because the redundant encoding introduced by the unified search space generally exist in high-dimensional problems. Noise encoding wastes many search resources in the search process and slows the overall progress. The separated encoding adopted by the MTEA-AST solves this problem well. In addition, compared to STO, MTEA-AST performs better on 11 and ties 9 out of 20 tasks in terms of the averaged objective value. MTEA-AST shows better performance on TSP and CVRP problems, while both have similar performance on QAP and LOP. On the one hand, the search space is larger in the benchmarks of TSP and CVRP. The prior knowledge introduced by the dimension unification strategy and the powerful local search ability of the task transfer strategy ensures that MTEA-AST can fully explore the search space. On the other hand, according to \cite{osabaATMFCGAAdaptiveTransferguided2021}, the similarity of instances in TSP and CVRP benchmarks is higher, and the transfer effect is more obvious. It is worth noting that because of the low similarity of instances in QAP and LOP benchmarks, the four compared MFO methods show obvious negative transfer phenomenon on all 20 tasks, while MTEA-AST only shows negative transfer in three benchmarks: ${{T}_{1}}$, ${{T}_{5}}$ of QAP and ${{T}_{5}}$ of LOP. It indicates that MTEA-AST can significantly suppress negative transfer in the MaTO environment with low similarity.

\subsubsection{Search efficiency}

\begin{figure*}[h] 
    \centering 
	%图2
	\subfloat[]{\includegraphics[height=3cm, width=\textwidth]{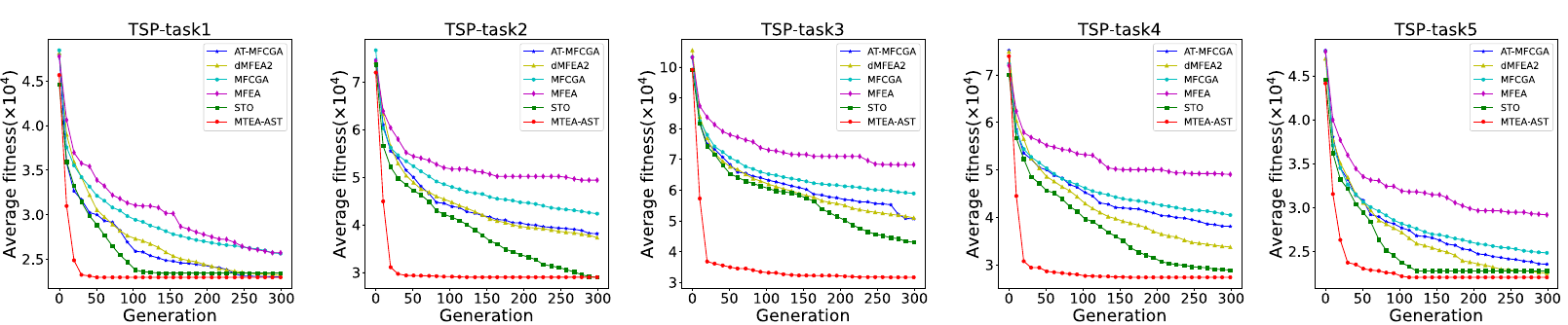}}\vspace{-0.4cm} \hfill 
	\subfloat[]{\includegraphics[height=3cm, width=\textwidth]{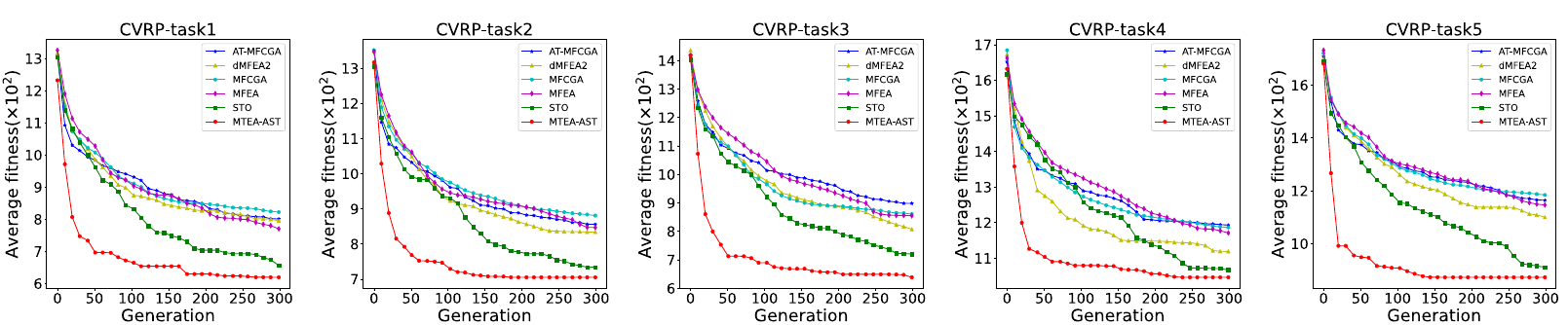}}\vspace{-0.4cm} \hfill 
	\subfloat[]{\includegraphics[height=3cm, width=\textwidth]{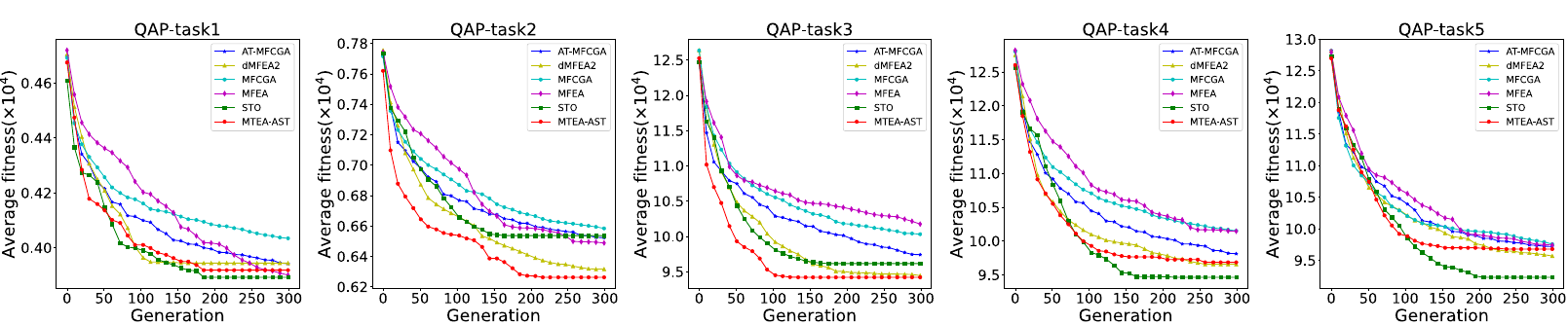}}\vspace{-0.4cm} \hfill 
	\subfloat[]{\includegraphics[height=3cm, width=\textwidth]{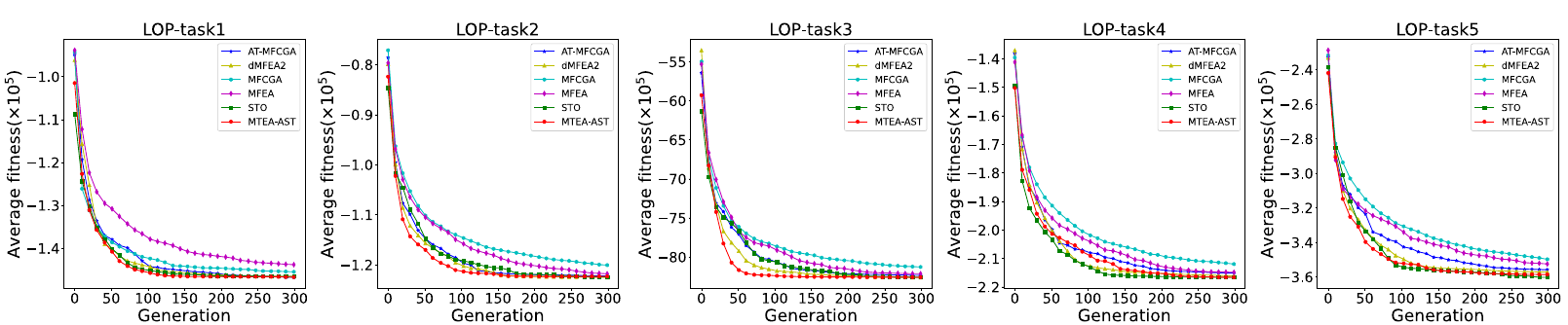}}
	\caption{Average convergence trend figures of all algorithms on 4 same-domain MaTO benchmarks.(a) TSP, (B) CVRP, (C) QAP, (D) LOP.\label{fig2}}
\end{figure*}

Fig. \ref{fig2} shows the average convergence trend of six methods on four same-domain MaTO benchmarks to illustrate the advantages of MTEA-AST in terms of search efficiency. The $x$-axis represents the number of algorithm iterations, and the $y$-axis represents the average objective function value of the population. According to Fig. \ref{fig2}, the search efficiency of MTEA-AST in the same-domain benchmarks is very competitive, especially in the TSP and CVRP. They have large search space and high similarity, so the powerful local search capability and excellent transfer effect of MTEA-AST speed up the search process, which is consistent with the conclusions in Table \ref{table2}. In QAP, limited by the low similarity between tasks, MTEA-AST also shows negative transfer. However, its transfer efficiency is still higher than the other four EMTO methods in the low similarity scenario. Moreover, in LOP, all tasks have the same dimension. Therefore, the search process of the MFO method is not affected by the redundant information of the unified search space, so the convergence trends of all methods are basically the same. In addition, we noticed that in scenarios with different task dimensions, the transfer effect of MTEA-AST is very significant in the early stage of high-dimensional task optimization. This indicates that our designed dimension unification strategy provides prior knowledge for each target task in the early stage of transfer through greedy heuristics. It makes the transfer effect in the early stage of optimization extremely significant, and the search process is accelerated by the guidance of prior knowledge.

\subsection{Results and discussion in cross-domain MaTO problems} \label{4.c}
This section shows the comparison results of solution quality and search efficiency between MTEA-AST
and other methods in cross-domain MaTO problems.
\subsubsection{Solution quality}

Table \ref{table3} shows the performance of MTEA-AST and five other compared methods on 7 cross-domain many-task benchmark problems. The first 6 test cases study the effect of cross-domain knowledge transfer between two types of tasks in MTEA-AST, and the last one studies the transfer effect between four types of tasks.

\begin{table*}[htp]
	\renewcommand{\arraystretch}{1.2}
	\caption{Comparative experimental results of all compared methods on benchmarks of cross-domain MaTO problems in terms of objective value (average ± standard deviation).\label{table3}}
	\centering
	\resizebox{0.9\textwidth}{!}{% 设置表格宽度
	\begin{tabular}{cccccccc}
		\hline
		Test case                   & Instance & MTEA-AST                      & STO                           & AT-MFCGA                & MFCGA                  & DMFEA-II                       & MFEA                    \\ \hline
		\multirow{10}{*}{TSP\_CVRP} & T1       & \textbf{22645.85±637.46}     & 22812.45±708.73($\approx$)           & 23329.13±559.3(-)       & 23621.87±918.14(-)     & 22882.96±617.81($\approx$)               & 25689.25±761.99(-)      \\
                            & T2       & \textbf{28491.68±509.35}     & 29197.06±635.1(-)            & 37235.38±1261.99(-)     & 38733.58±1729.24(-)    & 36012.53±1066.57(-)              & 43951.12±1224.22(-)     \\
                            & T3       & \textbf{31799.46±489.09}     & 40953.77±3291.4(-)           & 51685.51±1676.59(-)     & 53651.61±2069.69(-)    & 50635.82±1380.43(-)              & 61804.88±1279.99(-)     \\
                            & T4       & \textbf{27750.23±634.17}     & 28921.67±1455.28(-)          & 37160.31±992.93(-)      & 36431.64±2100.58(-)    & 35549.68±745.52(-)               & 42586.37±1151.9(-)      \\
                            & T5       & \textbf{22070.85±561.01}     & \textbf{21952.02±488.73($\approx$)}  & 23158.0±599.79(-)       & 23176.36±718.65(-)     & 22615.82±391.88(-)               & 25432.12±602.65(-)      \\
                            & T6       & \textbf{634.23±19.96}        & 671.48±42.45(-)              & 783.31±26.26(-)         & 792.09±24.97(-)        & 725.34±26.24(-)                  & 797.63±24.21(-)         \\
                            & T7       & \textbf{723.39±10.35}        & 755.43±30.74(-)              & 841.61±22.5(-)          & 861.56±29.94(-)        & 790.48±24.53(-)                  & 857.09±32.62(-)         \\
                            & T8       & \textbf{663.49±15.28}        & 699.9±30.06(-)               & 854.98±24.58(-)         & 881.45±28.02(-)        & 785.34±26.05(-)                  & 888.88±34.37(-)         \\
                            & T9       & \textbf{1034.25±20.23}       & 1085.39±26.0(-)              & 1174.38±27.82(-)        & 1194.81±20.02(-)       & 1118.8±25.74(-)                  & 1177.64±22.2(-)         \\
                            & T10      & \textbf{858.71±15.42}        & 949.83±27.43(-)              & 1097.97±33.51(-)        & 1130.16±42.3(-)        & 1024.36±33.89(-)                 & 1148.25±34.6(-)         \\ \hline
\multirow{10}{*}{TSP\_QAP}  & T1       & \textbf{22494.88±462.51}     & 22521.77±548.96($\approx$)           & 23641.92±692.55(-)      & 24040.11±720.14(-)     & \textbf{23136.92±611.0(-)}       & 25792.29±801.18(-)      \\
                            & T2       & \textbf{28437.53±448.1}      & 29479.33±765.4(-)            & 37514.61±1194.73(-)     & 37892.2±1418.71(-)     & \textbf{35961.95±1057.83(-)}     & 44269.68±1707.03(-)     \\
                            & T3       & \textbf{31889.54±814.52}     & 42799.66±2848.86(-)          & 52997.47±1311.24(-)     & 53253.15±2308.32(-)    & \textbf{50450.48±1128.88(-)}     & 61547.79±1306.44(-)     \\
                            & T4       & \textbf{27779.21±404.18}     & 29179.03±1213.45(-)          & 36855.97±882.31(-)      & 37386.55±1610.85(-)    & \textbf{35188.72±1070.05(-)}     & 42822.54±1364.57(-)     \\
                            & T5       & \textbf{22010.45±627.91}     & \textbf{22120.45±645.25($\approx$)}  & 23153.51±644.6(-)       & 23559.67±892.15(-)     & 22541.25±581.16(-)               & 25267.5±691.39(-)       \\
                            & T6       & \textbf{3855.3±46.3}         & 3874.3±47.85($\approx$)              & 3865.4±38.15($\approx$)         & 3879.9±29.05($\approx$)        & 3880.5±69.78($\approx$)                  & 3942.7±63.99(-)         \\
                            & T7       & \textbf{6345.0±60.65}        & 6360.1±85.11($\approx$)              & 6421.8±48.62(-)         & 6465.5±41.51(-)        & 6345.3±80.03($\approx$)                  & 6567.2±86.62(-)         \\
                            & T8       & \textbf{95315.0±1933.3}      & \textbf{94724.5±1479.22($\approx$)}  & 96130.5±1117.75($\approx$)      & 96453.5±1125.1($\approx$)      & 94743.5±2053.82($\approx$)               & 98289.5±1513.09(-)      \\
                            & T9       & \textbf{96058.0±1544.6}      & 95992.0±1918.32($\approx$)           & 96698.0±1079.19($\approx$)      & 96960.5±1277.6($\approx$)      & \textbf{95665.0±1195.12($\approx$)}      & 99576.0±1279.47(-)      \\
                            & T10      & 94546.0±1565.05              & \textbf{94538.0±1477.05($\approx$)}  & 95999.5±1176.36(-)      & 96413.0±1344.52(-)     & \textbf{94139.0±1577.07($\approx$)}      & 99756.5±1513.29(-)      \\ \hline
\multirow{10}{*}{TSP\_LOP}  & T1       & \textbf{22110.55±511.98}     & 22552.29±491.47(-)           & 23854.28±524.29(-)      & 23691.46±815.65(-)     & 23001.17±607.73(-)               & 26010.78±920.17(-)      \\
                            & T2       & \textbf{28143.52±529.38}     & 30035.13±980.6(-)            & 37719.04±1114.94(-)     & 38051.51±2077.0(-)     & 36045.09±953.16(-)               & 43761.9±1319.23(-)      \\
                            & T3       & \textbf{31629.45±499.84}     & 42428.32±1886.98(-)          & 52821.6±1367.41(-)      & 53653.01±2622.14(-)    & 50483.69±1158.92(-)              & 61711.78±1102.78(-)     \\
                            & T4       & \textbf{27537.06±334.52}     & 28895.66±984.93(-)           & 36642.48±1122.67(-)     & 37221.17±1600.31(-)    & 35272.33±1130.25(-)              & 42282.56±799.27(-)      \\
                            & T5       & \textbf{21677.67±504.27}     & 22169.61±471.67(-)           & 23223.59±439.24(-)      & 23623.84±594.12(-)     & 22687.92±437.66(-)               & 25391.07±872.02(-)      \\
                            & T6       & -146734.45±505.16            & -146768.15±357.22($\approx$)         & -146533.05±235.07($\approx$)    & -146517.0±237.71($\approx$)    & \textbf{-146832.35±402.58($\approx$)}    & -144988.4±778.02(-)     \\
                            & T7       & -122454.1±41.75              & -122354.2±240.96($\approx$)          & -121954.1±224.89(-)     & -121939.4±221.55(-)    & \textbf{-122477.75±53.52($\approx$)}     & -120962.2±662.74(-)     \\
                            & T8       & -8255409.45±14336.71         & -8246981.0±24283.9($\approx$)        & -8226467.0±15020.08(-)  & -8216518.4±16647.9(-)  & \textbf{-8257661.55±14637.17($\approx$)} & -8161001.25±33797.07(-) \\
                            & T9       & \textbf{-216660.85±279.97}   & -216732.15±145.05($\approx$)         & -214984.95±570.13(-)    & -215038.4±668.41(-)    & -216625.3±243.41($\approx$)              & -212623.15±1189.93(-)   \\
                            & T10      & \textbf{-359240.55±608.61}   & -359272.55±574.29($\approx$)         & -356173.05±920.83(-)    & -356032.9±1079.51(-)   & -358775.6±815.08(-)              & -352397.25±2407.65(-)   \\ \hline
\multirow{10}{*}{CVRP\_QAP} & T1       & \textbf{643.46±17.6}         & 670.19±22.92(-)              & 792.31±18.15(-)         & 818.34±21.09(-)        & 735.8±25.67(-)                   & 804.58±28.99(-)         \\
                            & T2       & \textbf{721.82±21.04}        & 755.52±22.19(-)              & 857.41±16.01(-)         & 870.78±30.71(-)        & 803.04±24.13(-)                  & 860.9±23.6(-)           \\
                            & T3       & \textbf{673.06±24.89}        & 716.44±34.04(-)              & 866.0±29.41(-)          & 896.07±30.55(-)        & 807.51±29.39(-)                  & 887.42±37.62(-)         \\
                            & T4       & \textbf{1046.84±18.34}       & 1093.61±24.41(-)             & 1179.79±17.37(-)        & 1199.7±27.0(-)         & 1140.62±26.09(-)                 & 1192.22±32.34(-)        \\
                            & T5       & \textbf{876.94±17.12}        & 935.34±33.59(-)              & 1132.05±27.16(-)        & 1139.74±26.83(-)       & 1051.06±32.98(-)                 & 1162.69±47.97(-)        \\
                            & T6       & 3862.7±53.33                 & 3859.5±68.37($\approx$)              & 3896.2±29.22(-)         & 3911.3±37.42(-)        & \textbf{3857.0±48.33($\approx$)}         & 3993.1±53.93(-)         \\
                            & T7       & \textbf{6362.1±56.26}        & 6384.5±86.95($\approx$)              & 6488.7±42.01(-)         & 6508.0±57.85(-)        & 6391.3±69.46($\approx$)                  & 6648.3±78.44(-)         \\
                            & T8       & 94994.5±1925.03              & 94816.5±1474.66($\approx$)           & 96596.0±1026.51(-)      & 97088.0±854.74(-)      & \textbf{94783.0±1936.41($\approx$)}      & 100603.0±2033.71(-)     \\
                            & T9       & 95915.0±1880.33              & 95858.0±2056.71($\approx$)           & 98042.5±1144.64(-)      & 98573.5±1133.19(-)     & \textbf{95655.0±1557.51($\approx$)}      & 101438.0±2230.54(-)     \\
                            & T10      & \textbf{94336.5±1736.66}     & 94410.0±1887.51($\approx$)           & 96960.0±1330.88(-)      & 97319.0±1207.03(-)     & 94604.0±1497.22($\approx$)               & 100955.5±1224.2(-)      \\ \hline
\multirow{10}{*}{CVRP\_LOP} & T1       & \textbf{643.23±15.49}        & 661.25±22.77(-)              & 799.3±18.56(-)          & 796.76±26.66(-)        & 739.21±24.93(-)                  & 810.32±22.46(-)         \\
                            & T2       & \textbf{715.24±13.52}        & 744.02±27.8(-)               & 855.51±21.26(-)         & 862.42±34.94(-)        & 807.76±24.61(-)                  & 877.98±22.74(-)         \\
                            & T3       & \textbf{664.09±22.55}        & 695.92±25.73(-)              & 866.83±23.75(-)         & 878.37±39.87(-)        & 821.59±25.98(-)                  & 893.13±22.66(-)         \\
                            & T4       & \textbf{1039.68±16.14}       & 1089.79±28.41(-)             & 1182.13±20.71(-)        & 1188.0±28.53(-)        & 1136.39±18.23(-)                 & 1192.26±26.1(-)         \\
                            & T5       & \textbf{855.28±19.69}        & 957.6±41.07(-)               & 1121.62±45.78(-)        & 1141.92±31.52(-)       & 1063.51±28.64(-)                 & 1156.08±25.52(-)        \\
                            & T6       & \textbf{-146764.05±338.62}   & -146639.15±452.87($\approx$)         & -146694.4±331.58($\approx$)     & -146494.4±320.25(-)    & -146647.5±478.25($\approx$)              & -144769.95±1051.44(-)   \\
                            & T7       & \textbf{-122433.15±155.37}   & -122382.4±130.15(-)          & -122044.3±178.98(-)     & -121870.5±284.6(-)     & -122415.35±106.62($\approx$)             & -120612.55±363.73(-)    \\
                            & T8       & \textbf{-8255991.1±14259.82} & -8240470.25±26981.51(-)      & -8231298.55±10027.79(-) & -8218657.1±10515.18(-) & -8251664.95±15543.73($\approx$)          & -8168220.35±32445.77(-) \\
                            & T9       & \textbf{-216741.5±240.96}    & -216699.25±273.73($\approx$)         & -215124.65±490.93(-)    & -215042.65±559.53(-)   & -216414.0±312.09(-)              & -212185.55±1110.29(-)   \\
                            & T10      & \textbf{-358824.3±668.74}    & -358821.7±788.41($\approx$)          & -356571.55±946.03(-)    & -355937.15±1084.17(-)  & -358222.4±828.62(-)              & -350508.0±2720.66(-)    \\ \hline
\multirow{10}{*}{QAP\_LOP}  & T1       & 3852.9±46.19                 & \textbf{3843.5±49.39($\approx$)}     & 3890.2±40.99(-)         & 3913.1±33.24(-)        & 3853.0±45.4($\approx$)                   & 4001.0±61.57(-)         \\
                            & T2       & \textbf{6332.0±84.42}        & 6371.3±62.58($\approx$)              & 6478.6±61.92(-)         & 6506.7±33.76(-)        & 6348.6±80.18($\approx$)                  & 6670.3±62.76(-)         \\
                            & T3       & \textbf{94541.0±1580.16}     & 95065.0±1910.27($\approx$)           & 97394.5±852.58(-)       & 97207.0±829.29(-)      & 94857.5±1413.06($\approx$)               & 101090.0±1746.2(-)      \\
                            & T4       & 95858.0±2000.05              & \textbf{95270.5±1307.99($\approx$)}  & 98176.0±1071.12(-)      & 97629.0±1219.07(-)     & 96254.0±1274.14($\approx$)               & 101632.0±1907.14(-)     \\
                            & T5       & 94736.5±1881.15              & \textbf{94289.5±1604.55($\approx$)}  & 96717.5±952.19(-)       & 97594.5±933.53(-)      & 94541.0±1549.78($\approx$)               & 100934.5±1162.41(-)     \\
                            & T6       & -146691.5±283.69             & -146620.6±437.7($\approx$)           & -146651.75±302.78($\approx$)    & -146483.7±391.84(-)    & \textbf{-146833.95±325.32($\approx$)}    & -144946.2±726.74(-)     \\
                            & T7       & \textbf{-122446.55±77.57}    & -122274.0±201.71(-)          & -122084.55±199.56(-)    & -122014.8±234.89(-)    & -122430.6±64.89($\approx$)               & -120722.35±653.59(-)    \\
                            & T8       & \textbf{-8259707.5±3948.52}  & -8237602.65±23909.61(-)      & -8231021.1±17845.72(-)  & -8223555.6±17593.65(-) & -8254664.95±7385.8(-)            & -8165555.05±26760.2(-)  \\
                            & T9       & \textbf{-216809.8±221.8}     & -216584.75±282.28(-)         & -215222.2±533.56(-)     & -215078.2±554.65(-)    & -216332.65±314.04(-)             & -212294.05±1384.18(-)   \\
                            & T10      & -359071.05±988.14            & \textbf{-359136.4±839.42(+)} & -356752.15±952.94(-)    & -356020.4±1286.75(-)   & -357975.4±1025.43(-)             & -350694.3±2179.81(-)    \\ \hline
\multirow{20}{*}{ALL}       & T1       & 22494.24±691.58              & \textbf{22389.1±360.35($\approx$)}   & 23912.8±625.72(-)       & 24278.47±816.93(-)     & 23347.76±415.16(-)               & 26503.99±548.98(-)      \\
                            & T2       & \textbf{28513.37±586.25}     & 29856.5±1113.74(-)           & 38152.95±1249.91(-)     & 38989.65±1257.4(-)     & 36769.09±974.89(-)               & 44247.65±1195.84(-)     \\
                            & T3       & \textbf{32101.48±744.48}     & 42254.67±3172.54(-)          & 53146.0±1572.31(-)      & 55303.75±1101.38(-)    & 50687.05±1304.59(-)              & 62566.19±1519.29(-)     \\
                            & T4       & \textbf{27855.65±485.87}     & 28950.37±1124.67(-)          & 38052.87±792.65(-)      & 38582.28±1375.12(-)    & 36010.52±844.8(-)                & 42955.93±1102.0(-)      \\
                            & T5       & \textbf{21965.16±487.01}     & 22062.94±695.51($\approx$)           & 23570.76±508.27(-)      & 23940.33±887.67(-)     & 22985.64±736.47(-)               & 25470.63±883.12(-)      \\
                            & T6       & \textbf{644.72±22.57}        & 666.71±26.29(-)              & 799.88±22.17(-)         & 809.95±28.21(-)        & 722.71±23.21(-)                  & 805.02±34.08(-)         \\
                            & T7       & \textbf{721.84±18.14}        & 743.14±26.36(-)              & 857.14±19.11(-)         & 875.07±26.77(-)        & 789.5±30.85(-)                   & 862.15±36.62(-)         \\
                            & T8       & \textbf{675.32±23.16}        & 717.32±27.9(-)               & 873.68±24.4(-)          & 890.5±24.28(-)         & 800.28±27.19(-)                  & 892.81±31.2(-)          \\
                            & T9       & \textbf{1048.72±15.15}       & 1092.21±39.16(-)             & 1190.28±22.97(-)        & 1199.92±27.66(-)       & 1123.82±20.77(-)                 & 1196.85±28.82(-)        \\
                            & T10      & \textbf{873.9±20.44}         & 953.91±34.97(-)              & 1119.96±28.18(-)        & 1147.92±46.18(-)       & 1026.2±29.45(-)                  & 1161.92±32.89(-)        \\
                            & T11      & 3866.9±49.54                 & 3871.4±53.63($\approx$)              & 3869.2±33.47($\approx$)         & 3875.2±38.57($\approx$)        & \textbf{3864.5±59.56($\approx$)}         & 3935.4±29.97(-)         \\
                            & T12      & 6364.4±77.69                 & \textbf{6352.7±62.93($\approx$)}     & 6448.9±51.92(-)         & 6469.3±34.83(-)        & 6355.8±76.46($\approx$)                  & 6553.1±100.04(-)        \\
                            & T13      & 94796.0±1864.8               & 94936.5±1619.89($\approx$)           & 96452.0±954.48(-)       & 97202.5±1210.54(-)     & \textbf{94171.5±1891.83($\approx$)}      & 98670.5±1702.44(-)      \\
                            & T14      & \textbf{95751.0±1715.56}     & 96116.5±1826.52($\approx$)           & 96942.5±958.61(-)       & 97247.0±958.31(-)      & 96351.5±1965.26($\approx$)               & 99836.5±1399.03(-)      \\
                            & T15      & \textbf{94213.5±1693.47}     & 94224.5±1524.42($\approx$)           & 96259.0±1092.32(-)      & 96687.0±1322.79(-)     & 94440.5±1645.03($\approx$)               & 98222.5±1896.32(-)      \\
                            & T16      & -146674.0±491.46             & -146731.75±486.43($\approx$)         & -146563.8±354.74($\approx$)     & -146360.15±318.56(-)   & \textbf{-146803.0±426.02($\approx$)}     & -145001.75±761.65(-)    \\
                            & T17      & -122451.4±64.68              & -122265.95±385.02(-)         & -121924.6±275.7(-)      & -121640.45±190.85(-)   & \textbf{-122470.55±60.12($\approx$)}     & -120779.5±664.0(-)      \\
                            & T18      & \textbf{-8259069.45±3953.07} & -8246699.3±19836.45(-)       & -8219994.1±15395.72(-)  & -8211003.1±20383.79(-) & -8254860.0±15181.47($\approx$)           & -8163211.8±27061.55(-)  \\
                            & T19      & \textbf{-216731.8±252.35}    & -216717.85±296.37($\approx$)         & -214847.5±594.53(-)     & -214640.45±452.39(-)   & -216593.1±255.77(-)              & -212218.6±1348.51(-)    \\
                            & T20      & \textbf{-359162.25±714.3}    & -358813.75±849.39($\approx$)         & -356447.6±765.69(-)     & -355434.5±1060.1(-)    & -358759.75±743.9($\approx$)              & -350861.2±2494.99(-)    \\ \hline
		\multicolumn{8}{l}{“$-$”, “$\approx$” and “+”   indicate that the corresponding compared method is significantly worse,   similar, and better than MTEA-AST, respectively.}                                                 
\end{tabular}}
\end{table*}

In addition to the larger number of tasks in the cross-domain benchmarks than in the same-domain benchmarks, the differences in dimension and solution representation between different tasks make the evolutionary environment more complex. According to Table \ref{table3}, MTEA-AST also outperforms AT-MFCGA, MFCGA, and MFEA in most 80 tasks on 7 cross-domain benchmarks. Specifically, compared to AT-MFCGA, MTEA-AST performers better on 72 and ties 8 out of 80 tasks in terms of the average objective value. Furthermore, it outperforms MFCGA on 76 tasks and ties on the remaining four tasks. In comparison with MFEA, MTEA-AST achieves competitive performance on all 80 tasks. The reasons for their poor performance in cross-domain transfer mainly come from two aspects. On the one hand, the dimensional differences in cross-domain benchmarks are much larger than in same-domain benchmarks. There is more redundant encoding in individuals responsible for low-dimensional tasks, resulting in a powerless search. On the other hand, the task selection mechanism of these frameworks is insensitive and cannot achieve fast capture of similarities between tasks and suppress negative transfer. This leads to a large number of negative transfer operations in the task relationship exploration stage, which affects the overall convergence speed. Especially, MFEA and MFCGA use the blind transfer strategy, which cannot dynamically capture the relationship between tasks and cannot suppress negative transfer. Moreover, in the comparison of MTEA-AST with DMFEA-II, the former performs better on 46, ties 34 out of 80 tasks. Specifically, the gap between them narrowed in almost all QAP tasks (TSP$\_$QAP ${{T}_{6}}\!\!\sim\!\!{{T}_{10}}$, CVRP ${{T}_{6}}\!\!\sim\!\!{{T}_{10}}$, QAP\_LOP ${{T}_{1}}\!\!\sim\!\!{{T}_{5}}$, ALL ${{T}_{11}}\!\!\sim\!\!{{T}_{15}}$) and some LOP tasks (TSP$\_$LOP ${{T}_{6}}\!\!\sim\!\!{{T}_{9}}$, CVRP\_LOP ${{T}_{6}}\!\!\sim\!\!{{T}_{8}}$, ALL ${{T}_{16}}\!\!\sim\!\!{{T}_{18}}$). This shows that DMFEA-II also performs well in cross-domain tasks, while MTEA-AST performs even better. Especially in higher-dimensional TSP and CVRP tasks, the dimension unification strategy and task transfer strategy play an extremely significant role in suppressing negative transfer and improving search efficiency. In addition, compared with STO, MTEA-AST performs better on 41, ties 38, and loss 1 out of 80 tasks. In 7 cross-domain benchmarks, MTEA-AST still has an advantage in TSP and CVRP. Similar to the results in the same-domain benchmarks, the performance of these two algorithms in QAP and LOP is basically equal. However, the performance of MTEA-AST on some TSP problems (${{T}_{1}}$, ${{T}_{5}}$) in TSP$\_$QAP and ALL also has a certain decline. This phenomenon shows that the transfer effect of MTEA-AST is also affected by the difficulty of cross-domain knowledge transfer, but the proposed strategy still works to suppress negative transfer.

\subsubsection{Search efficiency}

% 图4
\begin{figure*}[ht]
	\centering
	\includegraphics[width=\textwidth]{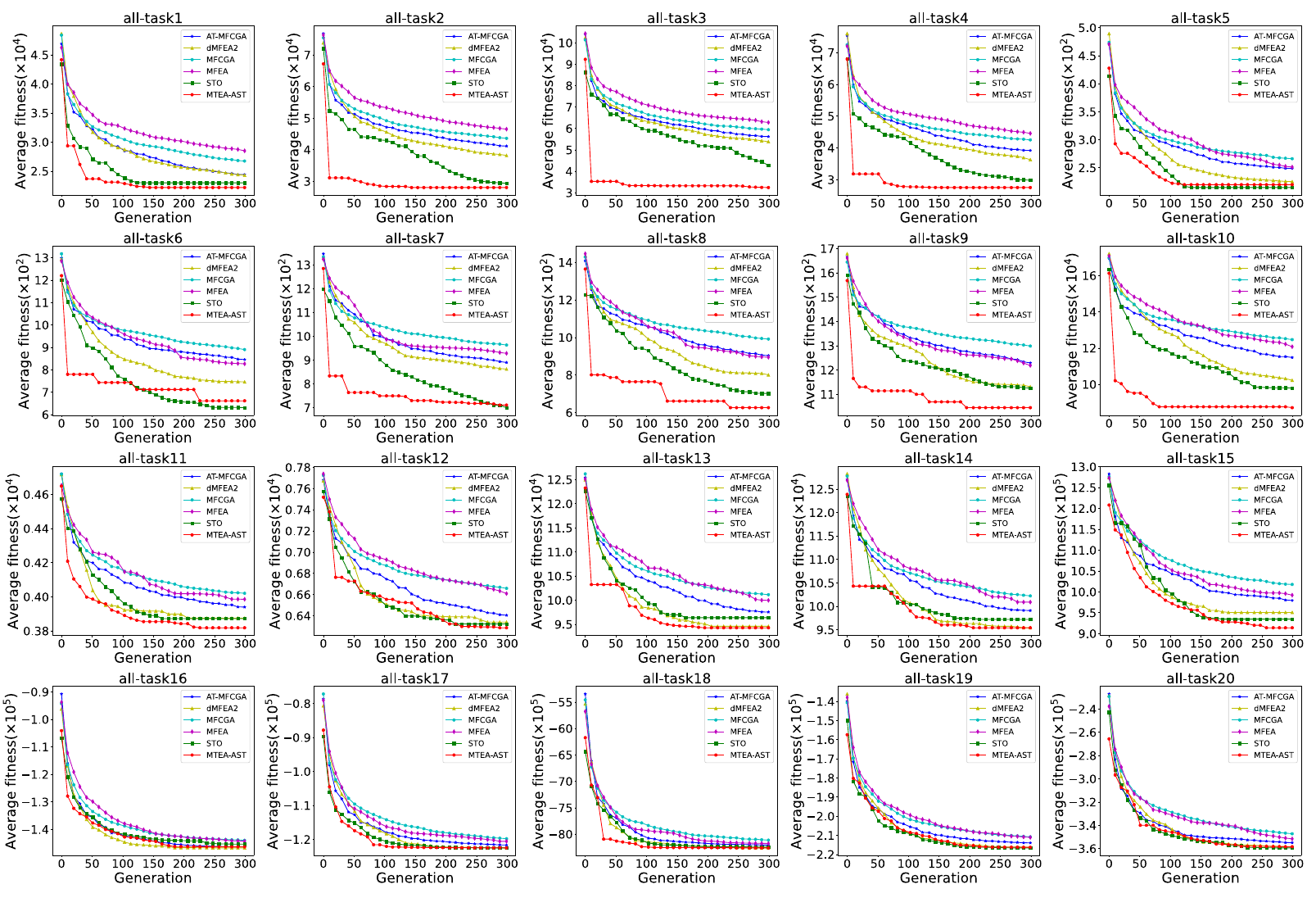}
	\caption{Average convergence trend figures of all algorithms on four-type cross-domain MaTO (ALL) benchmarks. \label{fig3}}
\end{figure*}

Fig. \ref{fig3} shows the average convergence trend of all algorithms on four-type cross-domain (ALL) benchmarks. According to Fig. \ref{fig3}, MTEA-AST outperforms the other five methods in search efficiency in TSP (${{T}_{1}}\!\!\sim\!\!{{T}_{5}}$), CVRP (${{T}_{6}}\!\!\sim\!\!{{T}_{10}}$) and QAP (${{T}_{11}}\!\!\sim\!\!{{T}_{15}}$). And the task dimensions in LOP (${{T}_{16}}\!\!\sim\!\!{{T}_{20}}$) are all equal, so the convergence trends of all algorithms are basically the same. This is similar to the experimental results in the same domain tasks. In cross-domain tasks, the performance of DMFEA-II and AT-MFCGA is much better than their behavior in same-domain tasks, especially in QAP and LOP. It shows that the dynamic task transfer strategies in MTEA-AST, DMFEA-II, and AT-MFCGA are more suitable for cross-domain MaTO tasks than blind static transfer strategies in MFCGA and MFEA. Concurrently, in QAP and LOP of cross-domain benchmarks, limited by the low similarity, the five EMTO methods show negative transfer. In addition, in the early stage of optimization of TSP, CVRP, and QAP, MTEA-AST still introduced prior knowledge to greatly speed up the optimization process by the dimensional unification strategy. Moreover, since the problem dimension gap in cross-domain tasks is larger than that in same-domain tasks, the advantages of MTEA-AST are more obvious in the early stage of optimization.

%---------------------------IV.	C.	Evaluation Metrics---------------------------------
\subsection{Effectiveness Analysis of Adaptive Task Selection Strategy} \label{4.d}

\begin{figure*}[htp] 
	\centering
	%图4
	\subfloat[]{\includegraphics[width=\textwidth]{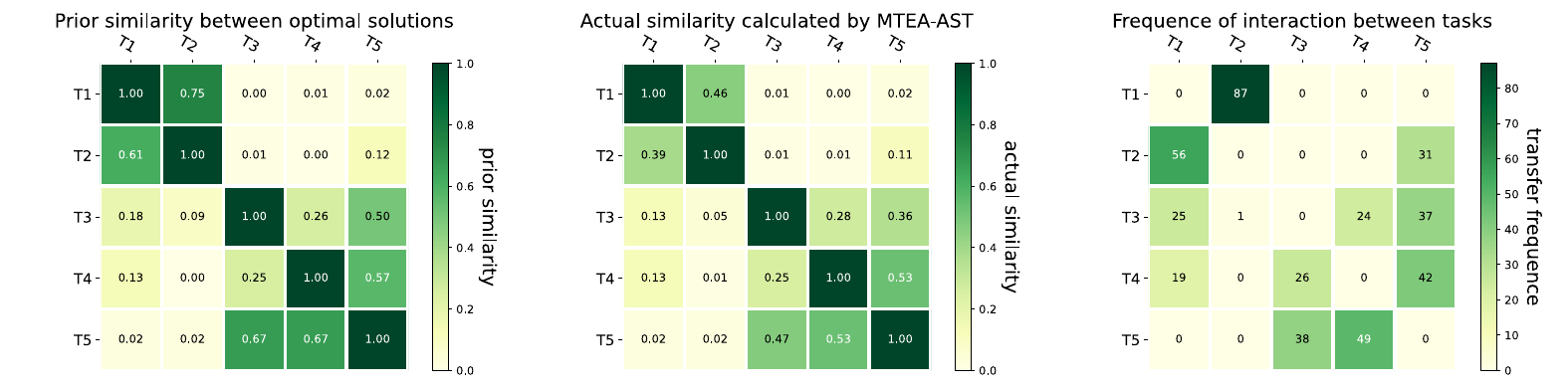}}\hfill
	\subfloat[]{\includegraphics[width=\textwidth]{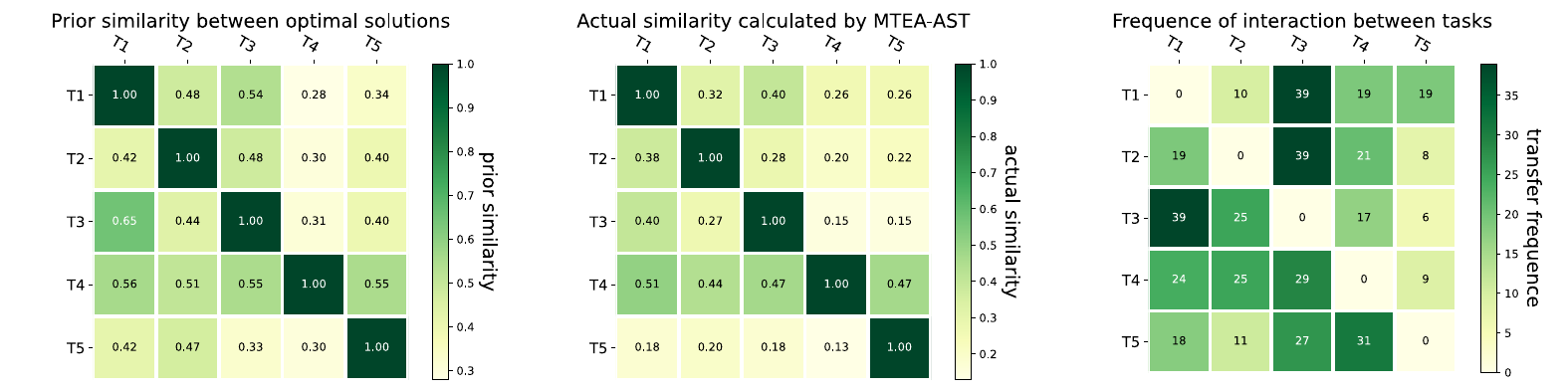}}\hfill
	\subfloat[]{\includegraphics[width=\textwidth]{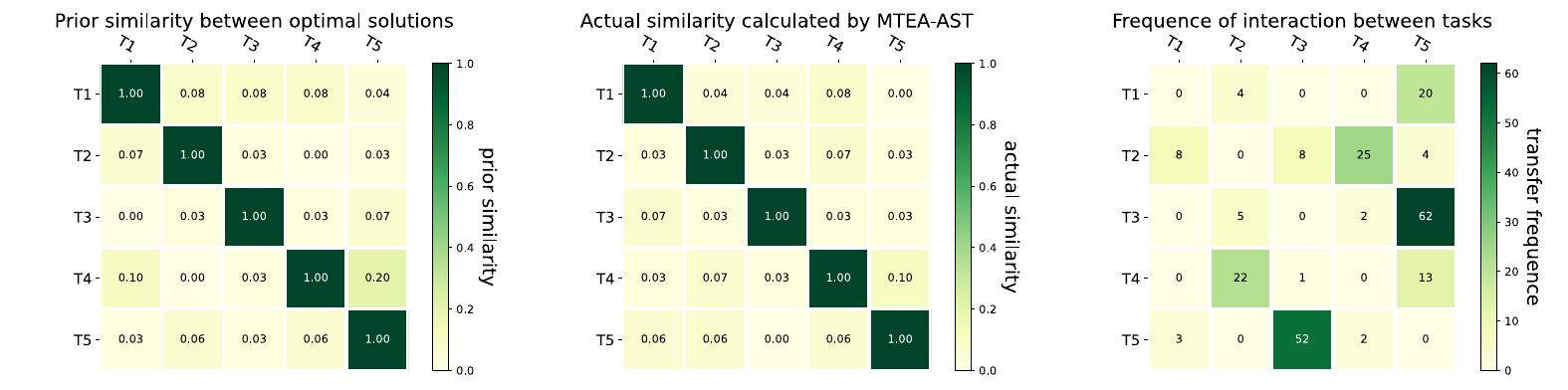}}\hfill
	\subfloat[]{\includegraphics[width=\textwidth]{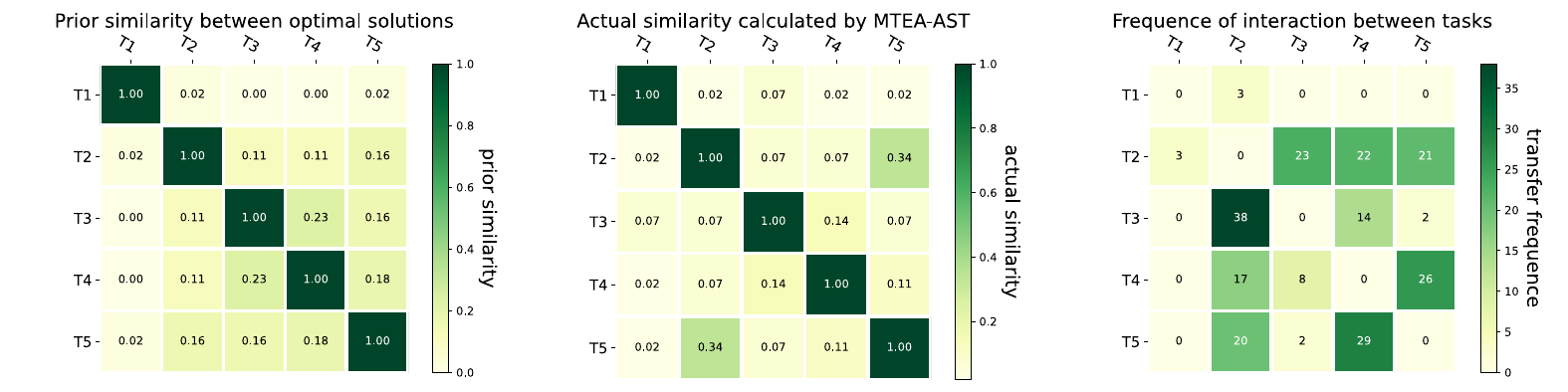}}
	\caption{Similarity and information exchange analysis in all same-domain problems; (a) TSP, (b) CVRP, (c) QAP, (d) LOP.\label{fig4}}
\end{figure*}

This section analyzes the effectiveness of the proposed adaptive task selection strategies in four same-domain benchmarks and some respective cross-domain benchmarks. In the subfigures of Fig. \ref{fig4}, the left subplot is the prior similarity of the global optimal solution for each task. The middle subplot is the similarity matrix obtained by the adaptive task selection strategy in MTEA-AST, i.e., the similarity between the best solutions of each population. The right subplot is the frequency of interactions between different tasks during the MTEA-AST run, i.e., the number of seeds transferred by each population for the target task. In addition, the $x$-axis of each heatmap is the source task, and the $y$-axis is the target task. Therefore, taking the interaction frequency heatmap in Fig. \ref{fig4}(a) as an example, ${{T}_{2}}$ provides 87 seeds to ${{T}_{1}}$, while ${{T}_{1}}$ provides 56 seeds to ${{T}_{2}}$.

\subsubsection{Same-domain problem}

According to Fig. \ref{fig4}(a), the high similarity of TSP tasks is mainly in two regions \{${{T}_{1}}$, ${{T}_{2}}$\} and \{${{T}_{3}}$, ${{T}_{4}}$, ${{T}_{5}}$\}. In the actual optimization process of MTEA-AST, the adaptive task selection strategy can capture the high similarity of these two regions. In addition, some other high similarities can also be obtained, such as ${{T}_{1}}$ between ${{T}_{3}}$ and ${{T}_{4}}$, ${{T}_{5}}$ between ${{T}_{2}}$. In terms of interaction frequency, the largest interaction frequency appears between \{${{T}_{1}}$, ${{T}_{2}}$\} with the highest similarity, and the distribution of high-frequency interactions is basically consistent with the similarity matrix. This illustrates that our adaptive task selection strategy performs well on TSP. In CVRP, the prior similarity of five tasks exceeds 10$\%$, where the similarity among \{${{T}_{1}}$, ${{T}_{2}}$, ${{T}_{3}}$\} and the similarity between ${{T}_{4}}$ and the other four tasks are all above 40$\%$. In Fig. \ref{fig4}(b), the figures of prior similarity and calculated similarity show the effectiveness of the adaptive task selection strategy in CVRP benchmarks. Meanwhile, the transfer frequency figure illustrates a high frequency of information exchange among the five CVRP tasks, more than 70 seeds are transferred to each target task (per row). Moreover, the high and uniform similarity distribution among tasks makes the transfer frequency distribution uniform. As shown in Fig. \ref{fig4}(c), the similarity among the five QAP tasks is mostly less than 10$\%$. In low similarity scenarios, our proposed adaptive task selection strategy still can capture the similarity of \{${{T}_{5}}$, ${{T}_{4}}$\}. And in terms of transfer frequency, high-frequency interactions randomly appear among some task pairs, but the total number of seeds transferred in each target task is significantly reduced. This shows that the total frequency of information exchange in low-similarity scenes is indeed reduced. However, in the early stage of optimization, the prior knowledge introduced by the dimension unification strategy and the seed growth strategy makes some seeds perform well in the target task population and become the best solution. Therefore, these seeds are used for similarity measure, which strengthens the interaction bias with the original responsible task. And a large number of seeds are migrated during the early stage of optimization. For LOP, the figure of prior similarity in Fig. \ref{fig4}(d) shows more than 10$\%$ similarity among the other four tasks except ${{T}_{1}}$, especially in \{${{T}_{3}}$, ${{T}_{4}}$\}, \{${{T}_{2}}$, ${{T}_{5}}$\}, \{${{T}_{4}}$, ${{T}_{5}}$\}. The similarity obtained by the adaptive task selection strategy agrees well with the prior similarity. And in terms of the transfer frequency, there is almost no information exchange between ${{T}_{1}}$ and the other four tasks, and the transfer frequency among the other four tasks is basically uniform. This is because the similarity between ${{T}_{1}}$ and the other four tasks is less than 10$\%$, and the adaptive task selection strategy prevents some information sharing between ${{T}_{1}}$ and other tasks through filters. Moreover, due to the low similarity, the transfer frequency between tasks is lower than that of TSP and CVRP. Another interesting phenomenon is that the number of seeds transferred from low-dimensional to high-dimensional tasks in the transfer frequency figures is greater than the opposite. For example, ${{T}_{1}}$ in Fig. \ref{fig4}(a) transferred significantly more seeds (${{T}_{1}}$ column) than ${{T}_{3}}$. The same phenomenon also exists in comparing ${{T}_{5}}$ with the other four tasks in CVRP benchmarks. However, LOPs with the same dimension have a uniform transfer frequency distribution. This is mainly due to the fact that the prior knowledge introduced by the dimension unification strategy for some individuals in the early stage of optimization shows advantageous fitness, and these individuals have a greater chance of being selected as seeds.

\subsubsection{Cross-domain problem}

\begin{figure*}[htp] 
	\centering
	%图5
	\subfloat[]{\includegraphics[width=\textwidth]{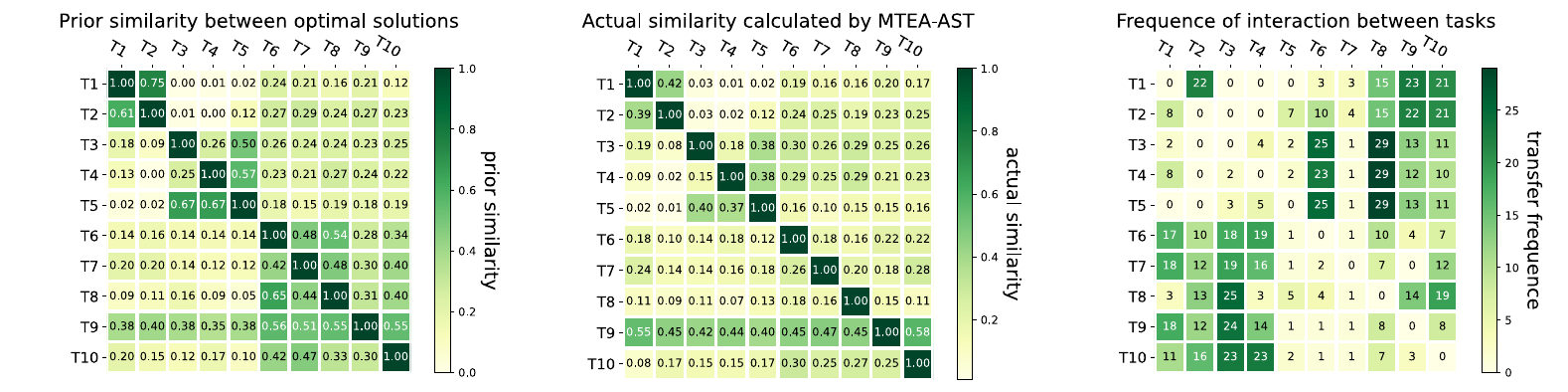}}\hfill
	\subfloat[]{\includegraphics[width=\textwidth]{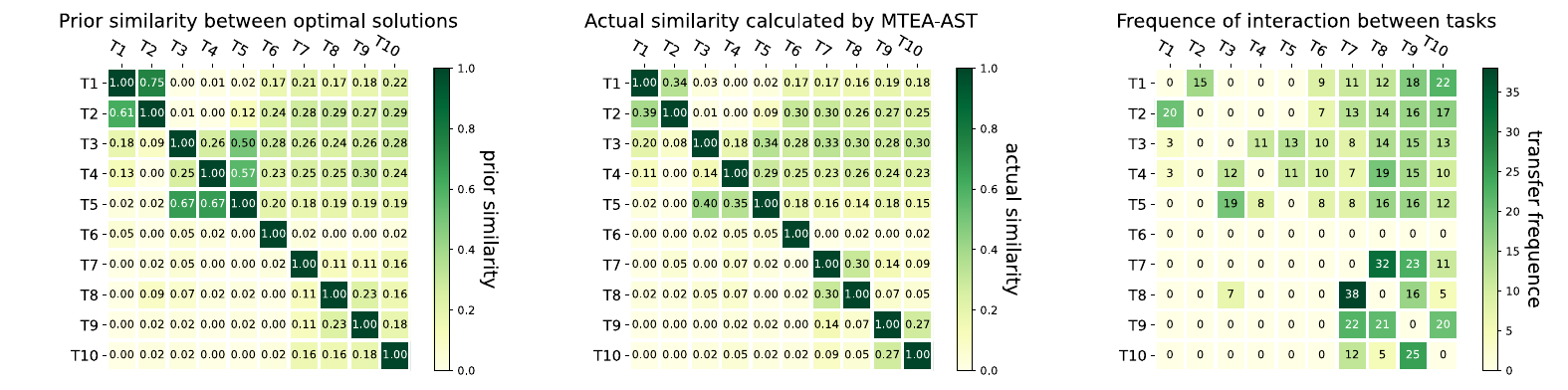}}\hfill
	\subfloat[]{\includegraphics[width=\textwidth]{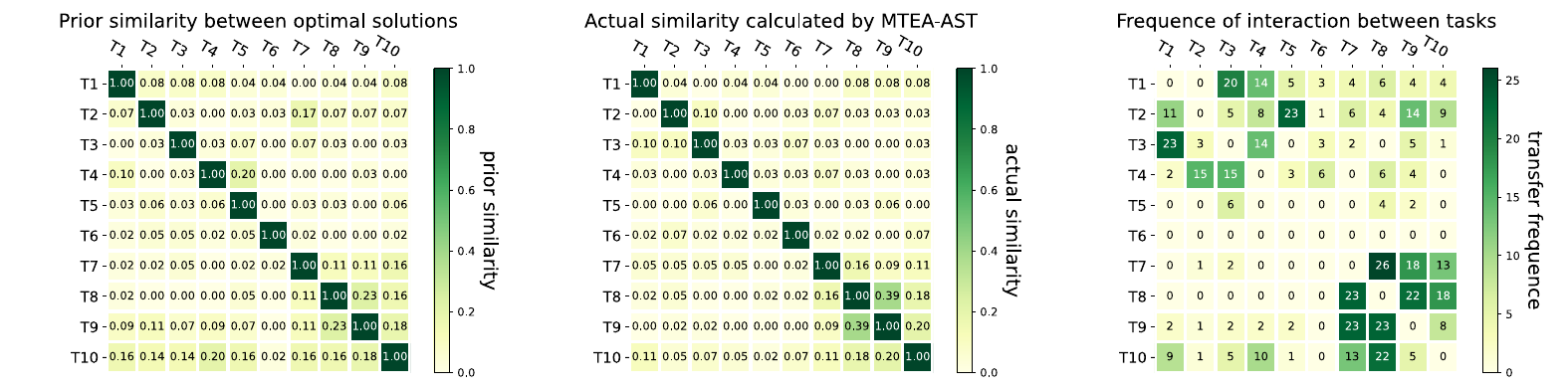}}\hfill
	\subfloat[]{\includegraphics[width=\textwidth]{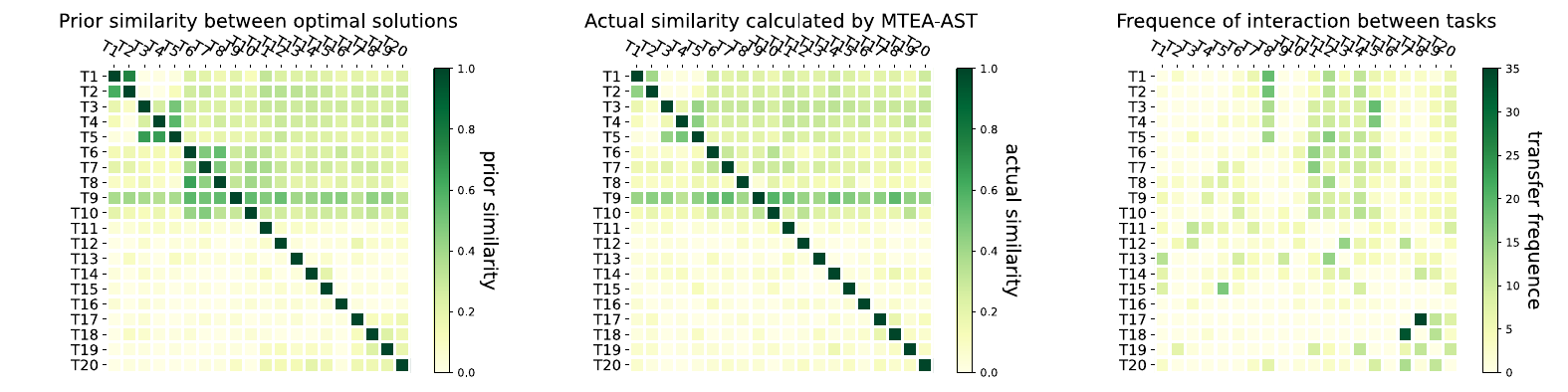}}
	\caption{Similarity and information exchange analysis in some respective cross-domain problems; (a) TSP$\_$CVRP, (b) TSP$\_$LOP, (c) QAP$\_$LOP, (d) ALL.\label{fig5}}
\end{figure*}

Fig. \ref{fig5} shows similarity and transfer frequency in several cross-domain benchmarks. According to Fig. \ref{fig5}(a), in the TSP\_CVRP benchmark, in addition to the same-domain similarity in these two types of problems, they also have high cross-domain similarity \{${T}_{1}\!\!\sim\!\!{T}_{5}\to {T}_{6}\!\!\sim\!\!{T}_{10}$\}\{${T}_{6}\sim{T}_{10}\to {T}_{1}\!\!\sim\!\!{T}_{5}$\}.The adaptive task selection strategy successfully captures their same-domain similarity, especially in TSP. The high similarity of CVRP is also captured, but it is less than in the same domain scene. And then, the cross-domain similarity of the two types of problems is fully obtained. It indicates that although the difficulty of cross-domain knowledge transfer affects the optimization effect to a certain extent, our proposed adaptive task selection strategy is feasible in capturing the same-domain task relationship as well as the cross-domain task relationship. The transfer frequency figure shows that there is frequent cross-domain information exchange between the two types of problems, which is closely related to their high cross-domain similarity. In Fig. \ref{fig5}(b), we observed that the cross-domain similarity between TSP and LOP is mainly reflected in \{${T}_{6}\!\!\sim\!\!{T}_{10}\to {T}_{1}\!\!\sim\!\!{T}_{5}$\}, that is, the one-way similarity between LOP and TSP. This phenomenon is mainly because of the similarity between the prior knowledge introduced by the dimension unification strategy and the optimal solution of TSP. In the similarity calculated in the actual optimization process, the same-domain similarity within the two types of problems has been successfully obtained \{${T}_{1}\!\!\sim\!\!{T}_{5}$\} \{${T}_{6}\!\!\sim\!\!{T}_{10}$\}. And the one-way similarity between the two types of problems is also accurately captured.The frequency of information exchange in the two types of problems is basically the same as that in the same-domain problems, and LOP also transfers a large number of seeds \{${T}_{6}\!\!\sim\!\!{T}_{10}\to {T}_{1}\!\!\sim\!\!{T}_{5}$\} to TSP due to their one-way similarity. Fig. \ref{fig5}(c) is the experimental result of the QAP\_LOP benchmark, the prior similarity shows the low cross-domain similarity between these two tasks. Therefore, in the actual optimization process, the adaptive task selection strategy only captures the same-domain similarity within each of the two types of tasks, and the same-domain similarity in LOP is basically the same as that in the same-domain task. In addition, in terms of transfer frequency, the same-domain transfer is much more than the cross-domain transfer. Finally, in Fig. \ref{fig5}(d), the similarity and transfer frequency among all tasks are discussed. The calculated actual similarity is very similar to the prior similarity, the high same-domain similarity in TSP, CVRP, and LOP is successfully demonstrated. The cross-domain similarity between different tasks is also captured. The cross-domain similarity is mainly shown in \{${T}_{6}\!\!\sim\!\!{T}_{20}\to {T}_{1}\!\!\sim\!\!{T}_{10}$\}, because compared with QAP \{${T}_{11}\!\!\sim\!\!{T}_{15}$\} and LOP \{${T}_{15}\!\!\sim\!\!{T}_{20}$\}, TSP \{${T}_{1}\!\!\sim\!\!{T}_{5}$\} and CVRP \{${T}_{6}\!\!\sim\!\!{T}_{10}$\} have higher dimensions, and the prior information introduced by the dimensional unification strategy to low-dimensional individuals increases the one-way similarity. Furthermore, the transfer frequency results also conform to the conclusions obtained earlier. CVRP and LOP maintain frequent same-domain information exchange with uniform and high same-domain similarity. And due to the low similarity between the task dimensions, QAP also transferred a large number of seeds to TSP and CVRP with the help of the same dimension strategy in the early stage of optimization.

%---------------------------IV.	D.	Experimental Details---------------------------------
\subsection{Effectiveness Analysis of Task transfer Strategy} \label{4.e}
To verify the effect of the designed task transfer strategy on suppressing negative transfer, we designed ablation experiments of MTEA-AST, STO, and MTEA-AST without task transfer strategy (MTEA-AST-noTS). In MTEA-AST-noTS, the number of selected individuals ${{p}_{i,j}}$ is still determined by the task adaptive selection strategy, and the first ${{p}_{i,j}}$ individuals with the best fitness value in the source task population are selected and directly inserted into the target task population. Table \ref{table 4} shows the average and standard deviation of three methods on ALL test cases. Three algorithms independently run 20 times, and a Wilcoxon rank-sum test with 95$\%$ confidence level has been conducted on the experimental results. According to the comparison results, compared to MTEA-AST-noTS, MTEA-AST performs better on 7 and ties 13 out of 20 tasks. compared to STO, MTEA-AST performs better on 10 and ties 10 out of 20 tasks. The difference between MTEA-AST-noTS and STO in TSP is mainly caused by the prior information introduced by the proposed dimension unification strategy, which greatly improves the optimization efficiency of the TSP problem in the early stage of optimization. In addition, in \{${{T}_{11}}\!\!\sim\!\!{{T}_{15}}$\} of QAP and \{${{T}_{16}}$,${{T}_{19}}$,${{T}_{20}}$\} of LOP, the performance of MTEA-AST-noTS is basically the same as or even slightly better than that of MTEA-AST. However, MTEA-AST-noTS performs slightly worse than STO and MTEA-AST on the high similarity CVRP problem. It shows that negative transfer is very common in the complex MaTO environment. The similarity in QAP and LOP benchmarks is very low, and the difference between MTEA-AST-noTS and MTEA-AST is not obvious. But MTEA-AST can fully suppress negative transfer to improve the utilization efficiency of task relationships in CVRP. It also confirms the important role of our designed task transfer strategy in suppressing negative transfer.

\begin{table}[htp]
	\renewcommand{\arraystretch}{1.2}
	\caption{Ablation experimental results of three methods on benchmarks of all test cases in terms of objective value (average ± standard deviation).\label{table 4}}
	\centering
        \scriptsize
	\setlength{\tabcolsep}{5.5mm}{
	% \resizebox{0.7\textwidth}{!}{ % 设置表格宽度
	\begin{tabular}{ccccc}
		\hline
		Test case                    & Instance       & MTEA-AST                           & MTEA-AST-noTS                      & STO                                 \\ \hline
		\multirow{20}{*}{ALL} & T1       & 22494.24±691.58              & 22662.38±577.89($\approx$)           & \textbf{22389.1±360.35($\approx$)} \\
                      & T2       & \textbf{28513.37±586.25}     & 28729.2±588.29($\approx$)            & 29856.5±1113.74(-)         \\
                      & T3       & \textbf{32101.48±744.48}     & 32579.98±786.63($\approx$)           & 42254.67±3172.54(-)        \\
                      & T4       & \textbf{27855.65±485.87}     & 28034.38±530.13($\approx$)           & 28950.37±1124.67(-)        \\
                      & T5       & \textbf{21965.16±487.01}     & 22085.51±553.75($\approx$)           & 22062.94±695.51($\approx$)         \\
                      & T6       & \textbf{644.72±22.57}        & 686.25±30.25(-)              & 666.71±26.29(-)            \\
                      & T7       & \textbf{721.84±18.14}        & 763.25±20.46(-)              & 743.14±26.36(-)            \\
                      & T8       & \textbf{675.32±23.16}        & 718.97±37.73(-)              & 717.32±27.9(-)             \\
                      & T9       & \textbf{1048.72±15.15}       & 1099.9±25.42(-)              & 1092.21±39.16(-)           \\
                      & T10      & \textbf{873.9±20.44}         & 945.08±37.7(-)               & 953.91±34.97(-)            \\
                      & T11      & \textbf{3866.9±49.54}        & 3869.0±55.27($\approx$)              & 3871.4±53.63($\approx$)            \\
                      & T12      & 6364.4±77.69                 & 6353.6±48.73($\approx$)              & \textbf{6352.7±62.93($\approx$)}   \\
                      & T13      & 94796.0±1864.8               & \textbf{94221.0±1528.67($\approx$)}  & 94936.5±1619.89($\approx$)         \\
                      & T14      & 95751.0±1715.56              & \textbf{95719.0±1585.46($\approx$)}  & 96116.5±1826.52($\approx$)         \\
                      & T15      & \textbf{94213.5±1693.47}     & 94533.0±1637.44($\approx$)           & 94224.5±1524.42($\approx$)         \\
                      & T16      & -146674.0±491.46             & \textbf{-146737.05±351.5($\approx$)} & -146731.75±486.43($\approx$)       \\
                      & T17      & \textbf{-122451.4±64.68}     & -122256.8±261.06(-)          & -122265.95±385.02(-)       \\
                      & T18      & \textbf{-8259069.45±3953.07} & -8251011.9±10628.95(-)       & -8246699.3±19836.45(-)     \\
                      & T19      & \textbf{-216731.8±252.35}    & -216651.8±249.56($\approx$)          & -216717.85±296.37($\approx$)       \\
                      & T20      & \textbf{-359162.25±714.3}    & -358959.3±699.84($\approx$)          & -358813.75±849.39($\approx$) \\ \hline
		\multicolumn{5}{l}{
			\begin{tabular}[c]{@{}l@{}}
			“$-$”, “$\approx$” and “+” indicate that the corresponding compared method is significantly worse, similar, \\and better than MTEA-AST, respectively.
		\end{tabular}
		}
	\end{tabular}}
\end{table}

\subsection{Analysis of the relationship between transfer effect and task similarity} \label{4.f}

\begin{figure}[h]
\begin{center}
\centerline{\includegraphics[width=0.57\linewidth]{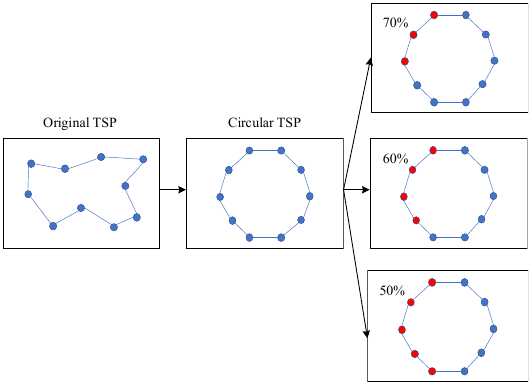}}
\caption{Schematic diagram of synthetic dataset construction. \label{fig7}}
\end{center}
\end{figure}

According to \cite{osabaMultifactorialCellularGenetic2020}, a similarity of 10$\%$ is sufficient to maintain minimal positive transfer effects. To further verify the relationship between similarity and transfer effect, we constructed task pairs with different prior similarities from the known optimal solutions of the TSP benchmark, and compared the performance of MTEA-AST and STO in these synthetic datasets. As is shown in Fig. \ref{fig7}, an existing TSP with a known global optimal solution is task 1, its global optimal solution is applied to construct a circular TSP task with the same best solution as task 1. After that, some nodes (red nodes) are chosen to scramble the sequence to obtain an auxiliary task (task 2), and its global optimal solution is only partially similar to that of task 1. Matching different auxiliary tasks with task 1 can get task pairs with different similarities. 

\begin{figure*}[h] 
	\centering
	%图5
	\subfloat[eil76]{\includegraphics[width=0.3\textwidth]{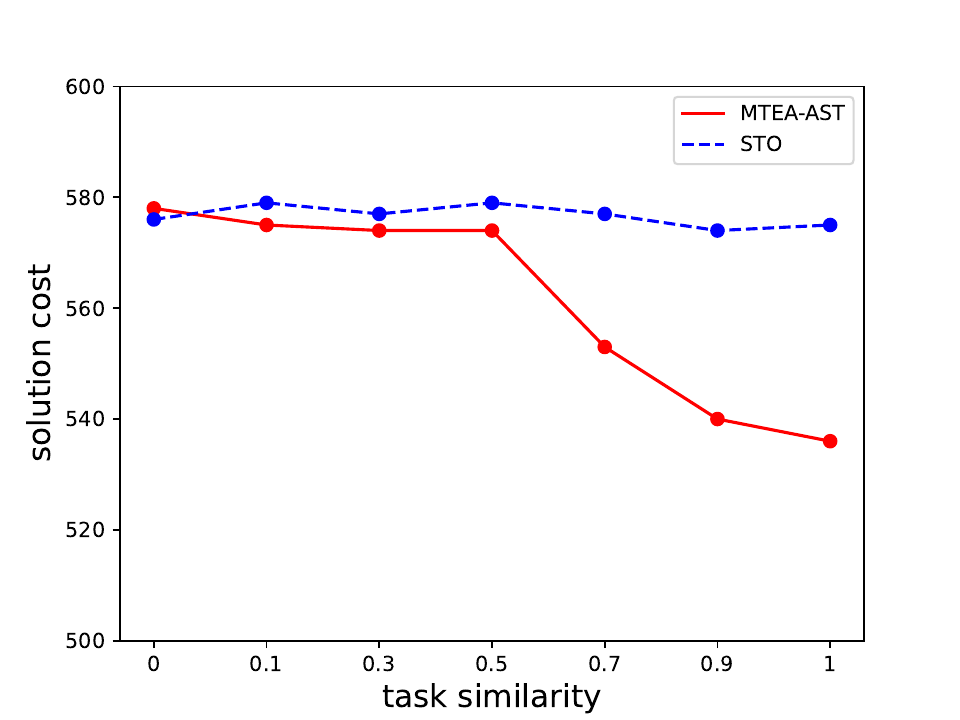}}\hfill
	\subfloat[kroA100]{\includegraphics[width=0.3\textwidth]{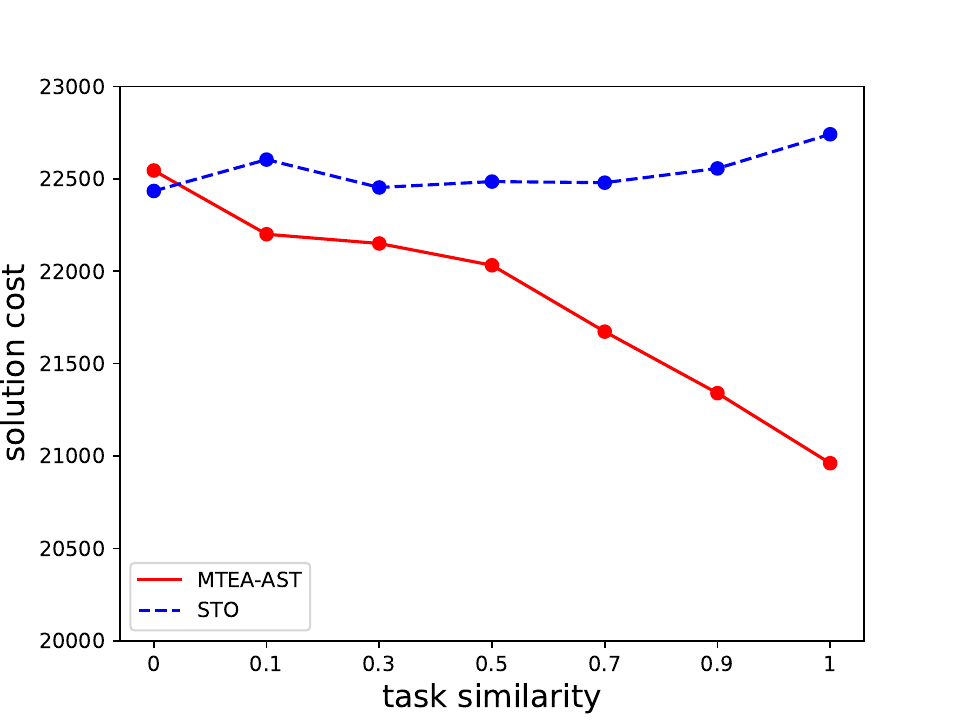}}\hfill
	\subfloat[ch130]{\includegraphics[width=0.3\textwidth]{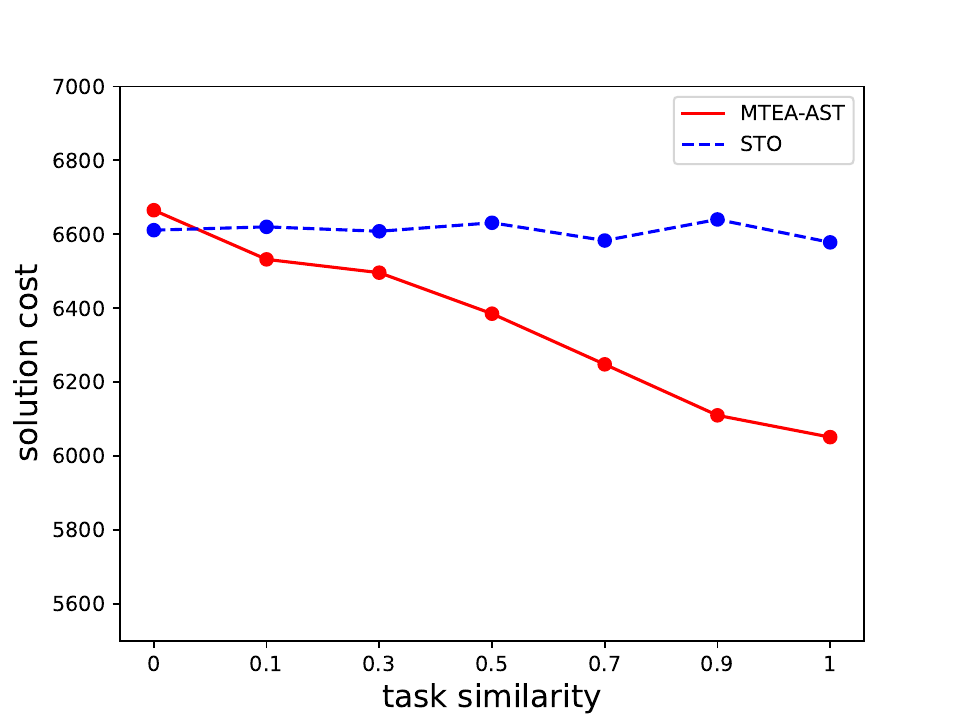}}\hfill
	\caption{Relationship between transfer effect and task similarity; (a) eil76, (b) kroA100, (c) ch130.\label{fig8}}
\end{figure*}

We make 21 test cases using three TSPLIB benchmarks (all of these benchmarks can be obtained in TSPLIB \footnote{http://comopt.ifi.uni-heidelberg.de/software/TSPLIB95/}) with different scales, and the similarity of each case is from 0 to 100$\%$. MTEA-AST and STO are used to perform optimization in these test cases, and the experimental results are shown in Fig. \ref{fig8}. MTEA-AST shows negative transfer in the test examples with 0 similarity in the three datasets, that is, the average value of solution cost is lower than that of STO. After that, MTEA-AST achieves similar performance to STO on the test examples with a similarity of 10$\%$ in the three datasets, even with a slight advantage. As the similarity increases, the positive transfer effect of MTEA-AST also expands. Therefore, in the proposed task adaptive selection operator, source tasks with a similarity less than 10$\%$ are screened out to avoid the transfer effect being affected by low-similarity source tasks.

%-------------------------------------------------------------V. Experimental Results------------------------------------------------
\section{Conclusion And Future Work}

In this work, a novel MTEA-AST method is proposed for multitasking COP. To address the dimensional inconsistency for multitasking COP, a new dimensional unification strategy is designed to unify the search space dimensions using greedy heuristics. It can also introduce prior knowledge to suppress negative transfer caused by the noise in the unified search space of MFO. Then, an adaptive task selection strategy is built to capture the relationships between tasks. The strategy calculates the similarity between tasks according to the Hamming distance of the current optimal solution of each task and dynamically determines its migration strength. In order to select more suitable seeds during migration and further suppress negative migration, a new transfer strategy involving both the seed selection mechanism and the seed growth mechanism was developed. Next, the performance of MTEA-AST is compared with five other state-of-the-art EMTO methods on 4 same-domain MaTO problems and 7 cross-domain MaTO problems. The comparative experimental results show the superiority of MTEA-AST. In addition, the effectiveness of the proposed adaptive task selection strategy is confirmed by analyzing the similarity and the frequency of information exchange between different tasks. Moreover, an ablation experiment in the full-task test case demonstrates the effectiveness of our proposed task transfer strategy for further suppressing negative transfer. Finally, we analyze the transfer effect of MTEA-AST between task similarity. In future work, it is planned to further improve and develop MTEA-AST from the following aspects. Developing seed transfer operators suitable for high-dimensional COPs; designing more efficient similarity measurement mechanisms for MaTO scenarios with more than a few hundred tasks; proposing corresponding individual mapping operators for more types of COPs.

%-------------------------------------------------------------Acknowledgment------------------------------------------------
\section*{Acknowledgment}
This work is supported in part by the Provincial Natural Science Foundation of Shaanxi of China (No. 2019JZ-26) and the National Natural Science Foundation of China (Nos. 61373111 and 61876141).

\bibliography{references}

\newpage

\appendix
\setcounter{table}{0}
\renewcommand\theequation{A.\arabic{table}}

\section{The transfer effect of all EMTO methods}

In the comparison experiment between the same-domain and cross-domain benchmarks, some EMTO methods show inferior performance than STO. In order to verify the effectiveness of the transfer operators of each EMTO method, we followed the dataset construction method in section \ref{4.f} and used TSPLIB datasets (all of these benchmarks can be obtained in TSPLIB \footnote{http://comopt.ifi.uni-heidelberg.de/software/TSPLIB95/}) with different scales to construct a series of synthetic task pairs with different similarities. Table \ref{tablea1} shows the comparison of the statistical results of all EMTO methods and STO on the synthetic dataset. First, both MFCGA and AT-MACGA show excellent performance in two datasets with lower dimensions (eil76 and pr76). In contrast to STO, they have positive transfer effects in test cases with all similarities of both datasets. This is mainly because the neighbor structure of the cellular genetic algorithm can provide a more stable local community structure compared to the random population structure, and can give full play to the advantages of excellent individuals to generate better offspring, which is very beneficial for exploring the search space of low-dimensional problems. The performance of MTEA-AST on both datasets is equally competitive. Although the task similarity of 10$\%$ cannot guarantee a huge positive transfer effect, it can already suppress the negative transfer with the help of this small amount of common knowledge, and achieve similar performance to STO. The task similarity of 30$\%$ can already guarantee the positive transfer effect of MTEA-AST. In contrast, the transfer effect of MFEA and DMFEA-II was much weaker. But in the 70$\%$ and 90$\%$ test examples, they can also show positive transfer effects, which shows that they are more suitable for high-similarity scenarios. 

In the test case of kroa100 and kroc100. MTEA-AST, MFCGA, and AT-MFCGA all showed negative transfer in the test samples with 0 similarities. At the same time, in the test example with a similarity of 10$\%$, all three have similar performance to STO, and MTEA-AST has a slight advantage. This is mainly due to the powerful local search capability of the MTEA-AST transfer operator. The similarity of 30$\%$ is enough to guarantee the positive transfer effect of these three methods on these two datasets. According to the performance of MFEA and DMFEA-II, their transfer operators are more suitable for task pairs with high similarity, and the positive transfer effect can only be achieved in the 90$\%$ test examples.

According to the performance of MTEA-AST in ch130 and ch150, the transfer operator performance of MTEA-AST is stable. In both datasets, 10$\%$ can still guarantee the lowest positive transfer effect to suppress negative transfer and 30$\%$ guarantees that it has a significant positive transfer effect. This is consistent with the conclusions we obtained in the first four datasets. The transfer effect of MFCGA and AT-MFCGA has declined in these two data sets. They are obviously suitable for high similarity scenarios in these two data sets, and the similarity is 70$\%$ and 90$\%$. The positive transfer effect is shown in the test samples. The decline in the effect of MFEA and DMFEA-II was even more severe, with negative transfer in all test cases. It shows that the transition operators in these two methods cannot adapt to the optimization environment of these two problems.

We can intuitively find that similarity and task dimension impact the transfer effect through the discussion of the above results. This is only in the task pair with the number of tasks being 2. In our studied scenario, the task selection mechanism in the MaTO environment and the dimensional difference and knowledge difference in the cross-domain transfer will make the actual optimization effect of the EMTO method worse than STO and lead to negative transfer. In our scenario, some tasks with high similarity mainly exist in the TSP and CVRP benchmarks. However, the dimensions of these two tasks are much larger than those of QAP and LOP. This leads to the poor transfer effect of the EMTO method, which is consistent with the experimental results in ch130 and ch150. The tasks in the synthetic dataset are all of the same dimension, and there are no redundant dimensions. The negative transfer caused by redundant dimensions between different tasks cannot be ignored in real scenarios. In the QAP and LOP benchmarks, the similarity between tasks is not high, not only in the other four EMTO methods, but also MTEA-AST also has a serious negative transfer phenomenon. This is consistent with our conclusion in all synthetic datasets, which shows that the transfer operator designed by the EMTO method cannot completely overcome the negative transfer between low-similarity tasks. Therefore, reasonable multi-task optimization modeling according to the real optimization requirements is also important to research content for the promotion of EMTO.

\begin{table}[htp]
  \centering
  \caption{Comparison of the statistical results  (average $\pm$ standard)  of all EMTO methods and STO obtained on synthetic datasets over 20 independent runs.\label{tablea1}}
\resizebox{0.95\textwidth}{!}{
% Please add the following required packages to your document preamble:
% \usepackage{multirow}
% \usepackage[table,xcdraw]{xcolor}
% If you use beamer only pass "xcolor=table" option, i.e. \documentclass[xcolor=table]{beamer}
\begin{tabular}{ccccccccc}
\hline
\multicolumn{1}{l}{dataset} & \multicolumn{1}{l}{similarity} & \multicolumn{1}{l}{task}  & MTEA-AST                                     & AT-MFCGA                                     & MFCGA                & DMFEA-II                                       & MFEA                 & STO               \\ \hline
                            &                                & T1                       & 578.69$\pm$7.76($\approx$)       & 567.34$\pm$5.9(+)        & 570.25$\pm$6.92(+)       & 584.36$\pm$12.9($\approx$)       & 586.68$\pm$8.28(-)       & 576.78$\pm$8.22       \\
                            & \multirow{-2}{*}{0}            & T2                       & 6.28$\pm$0.0($\approx$)          & 6.28$\pm$0.0(+)          & 6.28$\pm$0.0(+)          & 6.42$\pm$0.15(-)         & 6.37$\pm$0.12($\approx$)         & 6.29$\pm$0.04         \\
                            &                                & T1                       & 575.27$\pm$10.05($\approx$)      & 570.6$\pm$9.12(+)        & 569.86$\pm$5.36(+)       & 584.04$\pm$9.5($\approx$)        & 587.8$\pm$8.14(-)        & 579.21$\pm$10.25      \\
                            & \multirow{-2}{*}{0.1}          & T2                       & 6.29$\pm$0.04($\approx$)         & 6.28$\pm$0.0(+)          & 6.28$\pm$0.0(+)          & 6.41$\pm$0.15(-)         & 6.4$\pm$0.23($\approx$)          & 6.28$\pm$0.0          \\
                            &                                & T1                       & 577.5$\pm$7.99($\approx$)        & 567.16$\pm$4.79(+)       & 567.61$\pm$5.76(+)       & 583.49$\pm$8.62($\approx$)       & 582.85$\pm$11.62($\approx$)      & 577.82$\pm$8.08       \\
                            & \multirow{-2}{*}{0.3}          & T2                       & 6.25$\pm$0.04(+)         & 6.28$\pm$0.0(+)          & 6.28$\pm$0.0(+)          & 6.44$\pm$0.14(-)         & 6.41$\pm$0.13(-)         & 6.28$\pm$0.0          \\
                            &                                & T1                       & 574.11$\pm$8.85($\approx$)       & 566.55$\pm$6.02(+)       & 564.46$\pm$7.45(+)       & 581.73$\pm$8.89($\approx$)       & 584.17$\pm$11.06($\approx$)      & 579.72$\pm$10.77      \\
                            & \multirow{-2}{*}{0.5}          & T2                       & 6.21$\pm$0.03(+)         & 6.28$\pm$0.0(+)          & 6.28$\pm$0.0(+)          & 6.38$\pm$0.14($\approx$)         & 6.34$\pm$0.08(-)         & 6.28$\pm$0.0          \\
                            &                                & T1                       & 553.26$\pm$4.09(+)       & 553.35$\pm$5.91(+)       & 556.28$\pm$8.69(+)       & 579.38$\pm$11.2($\approx$)       & 577.43$\pm$7.69($\approx$)       & 577.43$\pm$10.01      \\
                            & \multirow{-2}{*}{0.7}          & T2                       & 6.2$\pm$0.0(+)           & 6.28$\pm$0.0(+)          & 6.28$\pm$0.0(+)          & 6.35$\pm$0.1(-)          & 6.33$\pm$0.09($\approx$)         & 6.3$\pm$0.07          \\
                            &                                & T1                       & 540.94$\pm$4.29(+)       & 545.39$\pm$0.0(+)        & 545.72$\pm$1.43(+)       & 556.92$\pm$13.41(+)      & 558.08$\pm$13.42(+)      & 574.86$\pm$6.79       \\
                            & \multirow{-2}{*}{0.9}          & T2                       & 6.2$\pm$0.0(+)           & 6.28$\pm$0.0(+)          & 6.28$\pm$0.0(+)          & 6.29$\pm$0.04($\approx$)         & 6.31$\pm$0.07($\approx$)         & 6.29$\pm$0.04         \\
                            &                                & T1                       & 536.47$\pm$2.37(+)       & 545.39$\pm$0.0(+)        & 545.39$\pm$0.0(+)        & 545.92$\pm$2.3(+)        & 547.31$\pm$5.61(+)       & 575.79$\pm$7.31       \\
\multirow{-14}{*}{eil76}    & \multirow{-2}{*}{1}            & T2                       & 6.2$\pm$0.0(+)           & 6.28$\pm$0.0(+)          & 6.28$\pm$0.0(+)          & 6.29$\pm$0.04($\approx$)         & 6.31$\pm$0.06($\approx$)         & 6.28$\pm$0.0          \\ \hline
                            &                                & T1                       & 112216.18$\pm$1353.01($\approx$) & 110560.08$\pm$1207.71(+) & 110064.06$\pm$965.7(+)   & 114684.24$\pm$2318.42(-) & 113874.02$\pm$2308.07(-) & 111612.39$\pm$1311.94 \\
                            & \multirow{-2}{*}{0}            & T2                       & 6.3$\pm$0.05($\approx$)          & 6.28$\pm$0.0(+)          & 6.28$\pm$0.0(+)          & 6.36$\pm$0.1(-)          & 6.34$\pm$0.09(-)         & 6.28$\pm$0.0          \\
                            &                                & T1                       & 111837.18$\pm$1797.32($\approx$) & 110949.76$\pm$1508.44($\approx$) & 110380.32$\pm$1074.68(+) & 113906.87$\pm$3353.66($\approx$) & 113634.93$\pm$1758.02(-) & 111765.1$\pm$1417.24  \\
                            & \multirow{-2}{*}{0.1}          & T2                       & 6.28$\pm$0.02($\approx$)         & 6.28$\pm$0.0(+)          & 6.28$\pm$0.0(+)          & 6.38$\pm$0.11(-)         & 6.36$\pm$0.12($\approx$)         & 6.32$\pm$0.15         \\
                            &                                & T1                       & 111168.96$\pm$1609.36($\approx$) & 110183.54$\pm$1651.54(+) & 110345.79$\pm$1364.98($\approx$) & 113532.07$\pm$1914.5(-)  & 113354.71$\pm$2697.18(-) & 111369.66$\pm$1638.39 \\
                            & \multirow{-2}{*}{0.3}          & T2                       & 6.24$\pm$0.04(+)         & 6.28$\pm$0.0(+)          & 6.28$\pm$0.0(+)          & 6.33$\pm$0.09($\approx$)         & 6.34$\pm$0.08($\approx$)         & 6.29$\pm$0.04         \\
                            &                                & T1                       & 110332.19$\pm$1797.68(+) & 109513.98$\pm$1130.25(+) & 110021.91$\pm$1674.06(+) & 112384.77$\pm$1753.06($\approx$) & 113735.05$\pm$2091.02(-) & 111919.98$\pm$1695.75 \\
                            & \multirow{-2}{*}{0.5}          & T2                       & 6.21$\pm$0.03(+)         & 6.28$\pm$0.0(+)          & 6.28$\pm$0.0(+)          & 6.39$\pm$0.14($\approx$)         & 6.37$\pm$0.13($\approx$)         & 6.29$\pm$0.04         \\
                            &                                & T1                       & 108257.57$\pm$1278.16(+) & 108502.38$\pm$949.64(+)  & 108264.74$\pm$459.0(+)   & 112249.49$\pm$3089.01($\approx$) & 112537.29$\pm$2494.42(-) & 111046.36$\pm$1731.93 \\
                            & \multirow{-2}{*}{0.7}          & T2                       & 6.2$\pm$0.0(+)           & 6.28$\pm$0.0(+)          & 6.28$\pm$0.0(+)          & 6.34$\pm$0.11($\approx$)         & 6.36$\pm$0.13(-)         & 6.28$\pm$0.0          \\
                            &                                & T1                       & 105946.65$\pm$1504.57(+) & 108159.44$\pm$0.0(+)     & 108159.44$\pm$0.0(+)     & 110017.42$\pm$2031.59(+) & 108925.0$\pm$1923.64(+)  & 111676.53$\pm$1717.41 \\
                            & \multirow{-2}{*}{0.9}          & T2                       & 6.2$\pm$0.0(+)           & 6.28$\pm$0.0(+)          & 6.28$\pm$0.0(+)          & 6.33$\pm$0.11($\approx$)         & 6.3$\pm$0.05($\approx$)          & 6.29$\pm$0.04         \\
                            &                                & T1                       & 105731.94$\pm$1004.79(+) & 108159.44$\pm$0.0(+)     & 108159.44$\pm$0.0(+)     & 108159.44$\pm$0.0(+)     & 108741.72$\pm$1264.96(+) & 111465.13$\pm$1772.18 \\
\multirow{-14}{*}{pr76}     & \multirow{-2}{*}{1}            & T2                       & 6.2$\pm$0.0(+)           & 6.28$\pm$0.0(+)          & 6.28$\pm$0.0(+)          & 6.28$\pm$0.0($\approx$)          & 6.32$\pm$0.13($\approx$)         & 6.28$\pm$0.0          \\ \hline
                            &                                & T1                       & 22545.38$\pm$622.39($\approx$)   & 22712.14$\pm$801.14($\approx$)   & 22361.2$\pm$499.65($\approx$)    & 24056.02$\pm$830.93(-)   & 24437.87$\pm$966.16(-)   & 22434.48$\pm$429.66   \\
                            & \multirow{-2}{*}{0}            & T2                       & 6.3$\pm$0.06($\approx$)          & 6.28$\pm$0.0($\approx$)          & 6.33$\pm$0.19($\approx$)         & 7.02$\pm$0.37(-)         & 7.3$\pm$1.51(-)          & 6.29$\pm$0.05         \\
                            &                                & T1                       & 22200.99$\pm$462.96(+)   & 22300.14$\pm$466.93($\approx$)   & 22369.41$\pm$485.55($\approx$)   & 24214.87$\pm$664.61(-)   & 24639.72$\pm$610.37(-)   & 22604.83$\pm$571.34   \\
                            & \multirow{-2}{*}{0.1}          & T2                       & 6.3$\pm$0.04($\approx$)          & 6.28$\pm$0.0(+)          & 6.28$\pm$0.0(+)          & 7.19$\pm$0.38(-)         & 7.07$\pm$0.44(-)         & 6.29$\pm$0.04         \\
                            &                                & T1                       & 22150.11$\pm$500.53($\approx$)   & 22327.3$\pm$538.22($\approx$)    & 22440.51$\pm$811.18($\approx$)   & 24385.85$\pm$652.09(-)   & 24261.41$\pm$994.05(-)   & 22453.61$\pm$646.46   \\
                            & \multirow{-2}{*}{0.3}          & T2                       & 6.27$\pm$0.02(+)         & 6.28$\pm$0.0(+)          & 6.28$\pm$0.0(+)          & 7.4$\pm$1.19(-)          & 7.26$\pm$1.25(-)         & 6.31$\pm$0.08         \\
                            &                                & T1                       & 22032.62$\pm$418.99(+)   & 21919.56$\pm$328.88(+)   & 21956.66$\pm$283.29(+)   & 24039.82$\pm$906.32(-)   & 23792.11$\pm$710.66(-)   & 22485.48$\pm$557.99   \\
                            & \multirow{-2}{*}{0.5}          & T2                       & 6.23$\pm$0.02(+)         & 6.28$\pm$0.0(+)          & 6.28$\pm$0.0(+)          & 7.07$\pm$0.38(-)         & 7.29$\pm$1.24(-)         & 6.28$\pm$0.0          \\
                            &                                & T1                       & 21673.17$\pm$198.72(+)   & 21531.27$\pm$221.9(+)    & 21666.25$\pm$240.99(+)   & 23762.4$\pm$1177.97(-)   & 23712.36$\pm$1032.13(-)  & 22479.79$\pm$647.42   \\
                            & \multirow{-2}{*}{0.7}          & T2                       & 6.22$\pm$0.0(+)          & 6.28$\pm$0.0(+)          & 6.28$\pm$0.0(+)          & 7.12$\pm$1.23(-)         & 7.29$\pm$1.23(-)         & 6.32$\pm$0.08         \\
                            &                                & T1                       & 21340.69$\pm$83.36(+)    & 21314.3$\pm$70.8(+)      & 21315.87$\pm$92.7(+)     & 23154.05$\pm$1186.23($\approx$)  & 22366.94$\pm$827.32($\approx$)   & 22556.72$\pm$639.41   \\
                            & \multirow{-2}{*}{0.9}          & T2                       & 6.22$\pm$0.0(+)          & 6.28$\pm$0.0(+)          & 6.28$\pm$0.0(+)          & 6.94$\pm$0.5(-)          & 7.04$\pm$1.31(-)         & 6.34$\pm$0.12         \\
                            &                                & T1                       & 20961.74$\pm$128.41(+)   & 21285.44$\pm$0.0(+)      & 21285.44$\pm$0.0(+)      & 21564.24$\pm$510.42(+)   & 21583.0$\pm$581.56(+)    & 22741.67$\pm$409.36   \\
\multirow{-14}{*}{kroa100}  & \multirow{-2}{*}{1}            & T2                       & 6.22$\pm$0.0(+)          & 6.28$\pm$0.0(+)          & 6.28$\pm$0.0(+)          & 6.43$\pm$0.26($\approx$)         & 6.78$\pm$1.35($\approx$)         & 6.31$\pm$0.07         \\ \hline
                            &                                & T1                       & 22239.08$\pm$461.53($\approx$)   & 22101.88$\pm$662.39($\approx$)   & 21941.91$\pm$523.48($\approx$)   & 24059.06$\pm$692.5(-)    & 24053.34$\pm$937.96(-)   & 21923.17$\pm$542.22   \\
                            & \multirow{-2}{*}{0}            & T2                       & 6.3$\pm$0.06($\approx$)          & 6.28$\pm$0.0($\approx$)          & 6.28$\pm$0.0(+)          & 7.86$\pm$1.94(-)         & 7.0$\pm$0.22(-)          & 6.34$\pm$0.11         \\
                            &                                & T1                       & 22165.23$\pm$495.28(-)   & 22056.81$\pm$691.93($\approx$)   & 22240.01$\pm$552.94(-)   & 23877.74$\pm$653.14(-)   & 23890.57$\pm$1009.94(-)  & 21884.06$\pm$437.8    \\
                            & \multirow{-2}{*}{0.1}          & T2                       & 6.29$\pm$0.03($\approx$)         & 6.28$\pm$0.0($\approx$)          & 6.29$\pm$0.03($\approx$)         & 7.18$\pm$0.36(-)         & 7.0$\pm$0.29(-)          & 6.31$\pm$0.06         \\
                            &                                & T1                       & 21709.11$\pm$511.07(+)   & 21763.61$\pm$646.22(+)   & 21843.36$\pm$463.47(+)   & 23652.07$\pm$695.51(-)   & 24071.67$\pm$827.53(-)   & 22191.9$\pm$603.56    \\
                            & \multirow{-2}{*}{0.3}          & T2                       & 6.24$\pm$0.03(+)         & 6.28$\pm$0.0($\approx$)          & 6.28$\pm$0.0(+)          & 7.3$\pm$1.21(-)          & 7.12$\pm$0.43(-)         & 6.35$\pm$0.15         \\
                            &                                & T1                       & 21283.83$\pm$404.98(+)   & 21367.89$\pm$375.76(+)   & 21459.16$\pm$616.75(+)   & 23396.13$\pm$1002.68(-)  & 23564.63$\pm$690.2(-)    & 22058.62$\pm$539.92   \\
                            & \multirow{-2}{*}{0.5}          & T2                       & 6.23$\pm$0.02(+)         & 6.28$\pm$0.0(+)          & 6.28$\pm$0.0(+)          & 7.22$\pm$1.2(-)          & 7.11$\pm$0.42(-)         & 6.33$\pm$0.1          \\
                            &                                & T1                       & 21171.76$\pm$423.57(+)   & 20980.71$\pm$345.0(+)    & 21003.52$\pm$326.91(+)   & 23079.12$\pm$821.57(-)   & 23258.59$\pm$1032.64(-)  & 22162.98$\pm$529.3    \\
                            & \multirow{-2}{*}{0.7}          & T2                       & 6.22$\pm$0.01(+)         & 6.28$\pm$0.0(+)          & 6.28$\pm$0.0(+)          & 6.97$\pm$0.38(-)         & 7.19$\pm$1.3(-)          & 6.33$\pm$0.09         \\
                            &                                & T1                       & 20678.49$\pm$157.15(+)   & 20797.9$\pm$50.17(+)     & 20782.66$\pm$76.45(+)    & 22405.62$\pm$867.92($\approx$)   & 22007.69$\pm$1086.76($\approx$)  & 22319.12$\pm$716.7    \\
                            & \multirow{-2}{*}{0.9}          & T2                       & 6.22$\pm$0.0(+)          & 6.28$\pm$0.0(+)          & 6.28$\pm$0.0(+)          & 6.9$\pm$0.42(-)          & 6.93$\pm$1.32(-)         & 6.32$\pm$0.08         \\
                            &                                & T1                       & 20464.49$\pm$105.88(+)   & 20750.76$\pm$0.0(+)      & 20750.76$\pm$0.0(+)      & 20879.71$\pm$369.45(+)   & 21140.22$\pm$721.56(+)   & 22066.36$\pm$499.13   \\
\multirow{-14}{*}{kroc100}  & \multirow{-2}{*}{1}            & T2                       & 6.22$\pm$0.0(+)          & 6.28$\pm$0.0(+)          & 6.28$\pm$0.0(+)          & 6.37$\pm$0.16($\approx$)         & 6.45$\pm$0.29($\approx$)         & 6.34$\pm$0.16         \\ \hline
                            &                                & T1                       & 6665.78$\pm$115.14($\approx$)    & 7159.16$\pm$341.47(-)    & 7080.83$\pm$380.76(-)    & 7756.65$\pm$265.2(-)     & 7708.55$\pm$273.94(-)    & 6611.78$\pm$152.92    \\
                            & \multirow{-2}{*}{0}            & T2                       & 7.48$\pm$2.39($\approx$)         & 6.99$\pm$1.94($\approx$)         & 6.96$\pm$2.0($\approx$)          & 10.7$\pm$2.1(-)          & 9.86$\pm$1.54(-)         & 6.6$\pm$1.27          \\
                            &                                & T1                       & 6532.14$\pm$144.55($\approx$)    & 7028.41$\pm$337.24(-)    & 7026.14$\pm$322.84(-)    & 7781.47$\pm$262.97(-)    & 7763.57$\pm$192.65(-)    & 6620.75$\pm$157.27    \\
                            & \multirow{-2}{*}{0.1}          & T2                       & 6.31$\pm$0.08($\approx$)         & 6.61$\pm$1.42($\approx$)         & 6.95$\pm$1.98($\approx$)         & 10.49$\pm$1.76(-)        & 10.1$\pm$1.73(-)         & 6.62$\pm$1.31         \\
                            &                                & T1                       & 6496.06$\pm$130.87(+)    & 7078.34$\pm$364.7(-)     & 7052.74$\pm$332.03(-)    & 7654.44$\pm$192.06(-)    & 7717.23$\pm$247.1(-)     & 6608.04$\pm$133.75    \\
                            & \multirow{-2}{*}{0.3}          & T2                       & 6.27$\pm$0.05(+)         & 6.61$\pm$1.32($\approx$)         & 6.69$\pm$1.5($\approx$)          & 10.56$\pm$1.94(-)        & 9.83$\pm$1.53(-)         & 6.32$\pm$0.11         \\
                            &                                & T1                       & 6385.06$\pm$125.83(+)    & 6797.78$\pm$258.12(-)    & 6797.76$\pm$317.46($\approx$)    & 7785.17$\pm$288.35(-)    & 7806.97$\pm$245.74(-)    & 6631.83$\pm$164.94    \\
                            & \multirow{-2}{*}{0.5}          & T2                       & 6.25$\pm$0.02(+)         & 6.39$\pm$0.38($\approx$)         & 6.3$\pm$0.06($\approx$)          & 10.75$\pm$2.15(-)        & 10.23$\pm$1.77(-)        & 6.62$\pm$1.32         \\
                            &                                & T1                       & 6248.37$\pm$70.88(+)     & 6382.31$\pm$173.89(+)    & 6476.01$\pm$204.87(+)    & 7705.02$\pm$239.82(-)    & 7671.16$\pm$352.51(-)    & 6583.76$\pm$121.02    \\
                            & \multirow{-2}{*}{0.7}          & T2                       & 6.24$\pm$0.01(+)         & 6.28$\pm$0.0($\approx$)          & 6.33$\pm$0.14($\approx$)         & 9.56$\pm$1.45(-)         & 9.82$\pm$1.91(-)         & 6.3$\pm$0.04          \\
                            &                                & T1                       & 6110.96$\pm$18.44(+)     & 6152.13$\pm$100.06(+)    & 6338.73$\pm$306.98(+)    & 7609.1$\pm$290.55(-)     & 7613.41$\pm$185.84(-)    & 6640.69$\pm$101.04    \\
                            & \multirow{-2}{*}{0.9}          & T2                       & 6.23$\pm$0.0(+)          & 6.29$\pm$0.04(+)         & 6.86$\pm$1.59($\approx$)         & 9.23$\pm$0.81(-)         & 10.37$\pm$1.82(-)        & 6.91$\pm$1.77         \\
                            &                                & T1                       & 6051.86$\pm$18.85(+)     & 6137.88$\pm$117.95(+)    & 6110.83$\pm$0.04(+)      & 7313.99$\pm$425.89(-)    & 7432.22$\pm$318.12(-)    & 6578.39$\pm$112.05    \\
\multirow{-14}{*}{ch130}    & \multirow{-2}{*}{1}            & T2                       & 6.23$\pm$0.0(+)          & 6.62$\pm$1.46($\approx$)         & 6.28$\pm$0.0($\approx$)          & 9.55$\pm$1.95(-)         & 9.69$\pm$1.56(-)         & 6.6$\pm$1.2           \\ \hline
                            &                                & T1                       & 7441.36$\pm$270.0($\approx$)     & 8684.84$\pm$295.8(-)     & 8606.3$\pm$459.8(-)      & 9279.37$\pm$265.12(-)    & 9283.22$\pm$285.37(-)    & 7360.83$\pm$225.99    \\
                            & \multirow{-2}{*}{0}            & T2                       & 7.46$\pm$2.31($\approx$)         & 7.47$\pm$2.43($\approx$)         & 7.06$\pm$1.08($\approx$)         & 12.2$\pm$2.02(-)         & 12.87$\pm$1.95(-)        & 7.93$\pm$2.78         \\
                            &                                & T1                       & 7464.91$\pm$242.15($\approx$)    & 8468.34$\pm$495.81(-)    & 8442.05$\pm$460.42(-)    & 9289.29$\pm$251.18(-)    & 9260.65$\pm$265.87(-)    & 7460.53$\pm$317.62    \\
                            & \multirow{-2}{*}{0.1}          & T2                       & 6.63$\pm$1.45($\approx$)         & 6.67$\pm$0.63($\approx$)         & 7.34$\pm$1.91(-)         & 12.32$\pm$1.81(-)        & 12.31$\pm$1.84(-)        & 7.31$\pm$2.4          \\
                            &                                & T1                       & 7191.65$\pm$305.78(+)    & 8552.26$\pm$344.65(-)    & 8428.86$\pm$515.84(-)    & 9137.02$\pm$278.29(-)    & 9165.73$\pm$287.34(-)    & 7443.62$\pm$233.92    \\
                            & \multirow{-2}{*}{0.3}          & T2                       & 6.27$\pm$0.02(+)         & 7.27$\pm$2.09($\approx$)         & 7.17$\pm$0.69(-)         & 12.33$\pm$2.09(-)        & 12.0$\pm$1.81(-)         & 6.45$\pm$0.6          \\
                            &                                & T1                       & 6882.85$\pm$145.11(+)    & 8220.48$\pm$363.43(-)    & 8167.8$\pm$593.38(-)     & 9199.97$\pm$224.16(-)    & 9188.7$\pm$231.71(-)     & 7371.12$\pm$265.59    \\
                            & \multirow{-2}{*}{0.5}          & T2                       & 6.25$\pm$0.02(+)         & 8.25$\pm$2.89($\approx$)         & 8.27$\pm$2.88($\approx$)         & 12.55$\pm$1.46(-)        & 12.35$\pm$1.86(-)        & 7.88$\pm$2.72         \\
                            &                                & T1                       & 6694.64$\pm$86.6(+)      & 7786.06$\pm$459.7($\approx$)     & 7755.71$\pm$645.31($\approx$)    & 9113.63$\pm$309.84(-)    & 9262.77$\pm$295.79(-)    & 7519.58$\pm$280.38    \\
                            & \multirow{-2}{*}{0.7}          & T2                       & 6.24$\pm$0.01(+)         & 7.24$\pm$1.63($\approx$)         & 7.91$\pm$2.88($\approx$)         & 12.71$\pm$1.74(-)        & 12.45$\pm$1.84(-)        & 8.62$\pm$3.17         \\
                            &                                & T1                       & 6566.36$\pm$48.25(+)     & 6861.99$\pm$513.91(+)    & 7026.44$\pm$758.13(+)    & 9214.51$\pm$279.01(-)    & 9061.23$\pm$254.73(-)    & 7439.29$\pm$249.48    \\
                            & \multirow{-2}{*}{0.9}          & T2                       & 6.24$\pm$0.01(+)         & 6.58$\pm$0.63($\approx$)         & 7.47$\pm$2.28($\approx$)         & 12.66$\pm$2.21(-)        & 11.96$\pm$1.9(-)         & 7.42$\pm$2.59         \\
                            &                                & T1                       & 6468.67$\pm$18.77(+)     & 6731.47$\pm$491.58(+)    & 6888.93$\pm$589.03(+)    & 9063.47$\pm$255.2(-)     & 9087.7$\pm$216.4(-)      & 7415.52$\pm$243.36    \\
\multirow{-14}{*}{ch150}    & \multirow{-2}{*}{1}            & T2                       & 6.24$\pm$0.0(+)          & 6.86$\pm$1.78($\approx$)         & 7.15$\pm$1.89($\approx$)         & 12.14$\pm$1.68(-)        & 12.1$\pm$1.7(-)          & 7.94$\pm$2.8  \\ \hline       
\end{tabular}
}
\end{table}

\end{document}